\documentclass[runningheads]{llncs}
\usepackage{graphicx}
\usepackage{comment}
\usepackage{amsmath,amssymb} 
\usepackage{xcolor}
\usepackage{subfig}
\usepackage{bm}
\usepackage{hyperref}
\hypersetup{
    colorlinks=true,
    linkcolor=blue,
    filecolor=magenta,      
    urlcolor=blue,
}
\usepackage[figuresleft]{rotating}
\usepackage{textcomp}
\usepackage{makecell}

\usepackage{soul}

\newcommand\blfootnote[1]{%
  \begingroup
  \renewcommand\thefootnote{}\footnote{#1}%
  \addtocounter{footnote}{-1}%
  \endgroup
}

\def\eg{\emph{e.g. }}
\def\ie{\emph{i.e. }}
\def\etal{\emph{et al. }}

\renewcommand\v[1]{\mathbf{#1}}
\newcommand\curframe{\v{x}_t}
\newcommand\prevframe{\v{x}_{t-1}}

\newcommand\prevrecon{\widehat{\v{x}}_{t-1}}

\newcommand{\menet}{\operatorname{MENet}}
\newcommand{\warp}{\texttt{warp}}

\usepackage[normalem]{ulem}

\usepackage[width=122mm,left=12mm,paperwidth=146mm,height=193mm,top=12mm,paperheight=217mm]{geometry}

\begin{document}

\pagestyle{headings}
\mainmatter
\def\ECCVSubNumber{5911}  

\title{Feedback Recurrent Autoencoder for\\Video Compression}

\titlerunning{ }

%

\author{Adam Goli\'nski\inst{*\dagger3} \and
Reza Pourreza\inst{*1} \and
Yang Yang\inst{*1} \and 
Guillaume Sauti\`ere\inst{2} \and
Taco S Cohen\inst{2}
}
%
\authorrunning{ }

%
\institute{Qualcomm AI Research, Qualcomm Technologies, Inc.\\ \and
Qualcomm AI Research, Qualcomm Technologies Netherlands B.V.\\ \and
Department of Engineering Science, University of Oxford
}

\maketitle
\begin{abstract}

Recent advances in deep generative modeling have enabled efficient modeling of high dimensional data distributions and opened up a new horizon for solving data compression problems. Specifically, autoencoder based learned image or video compression solutions are emerging as strong competitors to traditional approaches. In this work, We propose a new network architecture, based on common and well studied components, for learned video compression operating in low latency mode. Our method yields state of the art MS-SSIM/rate performance on the high-resolution UVG dataset, among both learned video compression approaches and classical video compression methods (H.265 and H.264) in the rate range of interest for streaming applications. Additionally, we provide an analysis of existing approaches through the lens of their underlying probabilistic graphical models.
Finally, we point out issues with temporal consistency and color shift observed in empirical evaluation, and suggest directions forward to alleviate those.

\end{abstract}
\blfootnote{$^*$ Equal contribution\\$^\dagger$ Work completed during internship at Qualcomm Technologies Netherlands B.V.\\Qualcomm AI Research is an initiative of Qualcomm Technologies, Inc.}
\vspace{-1cm}

\section{Introduction}

With over 60\% of internet traffic consisting of video \cite{video-data-global-usage}, lossy video compression is a critically important problem, promising reductions in bandwidth, storage, and generally increasing the scalability of the internet.
Although the relation between probabilistic modelling and compression has been known since Shannon, video codecs in use today are 
only to a very small extent
based on learning and are 
not end-to-end optimized for rate-distortion performance on a large and representative video dataset.

The last few years have seen a surge in interest in novel codec designs based on deep learning \cite{luDVCEndtoendDeep2018,wuVideoCompressionImage2018,rippelLearnedVideoCompression2018,Habibian_2019_ICCV,liu2020learned,stephan_2019_neurips}, which are trained end-to-end to optimize rate-distortion performance on a large video dataset, and
a large number of network designs with various novel,
often domain-specific
components have been proposed.
In this paper we show that a relatively simple design based on standard, well understood and highly optimized components such as residual blocks \cite{heDeepResidualLearning2016}, convolutional recurrent networks \cite{convlstm}, optical flow warping and a PixelCNN \cite{vandenOord2016pixelCNN} prior yields state of the art rate-distortion performance.

We focus on the online compression setting, where video frames can be compressed and transmitted as soon as they are recorded (in contrast to approaches which require buffering several frames before encoding), which is necessary for applications such as video conferencing and cloud gaming. 
Additionally, in both applications, the ability to finetune the neural codec to its specific content
holds promise to further significantly reduce the required bandwidth \cite{Habibian_2019_ICCV}.

There are two key components to our approach beyond the residual block based encoder and decoder architecture. 
Firstly, to exploit long range temporal correlation, we follow the principle proposed in 
Feedback Recurrent AutoEncoder (FRAE) \cite{FRAE}, which is shown to be effective for speech compression, by adding a convolutional GRU module in the decoder and feeding back the recurrent state to the encoder. 
Secondly,we apply a motion estimation network at the encoder side and enforce optical flow learning by using an explicit loss term during the initial stage of the training, which leads to better optical flow output at the decoder side and consequently much better rate-distortion performance. 
The proposed network architecture is compared with existing learned approaches through the lens of their underlying probabilistic models in Section~\ref{sec:related_work}.

We compare our method with state-of-the-art traditional codecs and learned approaches on the high resolution UVG \cite{UVG} video dataset by plotting rate-distortion curves, with MS-SSIM \cite{wangMultiScaleStructuralSimilarity2003} as a distortion metric. We show that we outperform all methods in the low to high rate regime, and more specifically in the 0.09-0.13 bits per-pixel (bpp) region, which is of practical interest for video streaming \cite{netflix-data-usage}.

To summarize, our main contributions are as follows:
\begin{enumerate}
    \item We develop a simple feedback recurrent video compression architecture based on widely used building blocks. 
    (Section \ref{sec:methodology}).
    \item We study the differences and connections of existing learned video compression methods by detailing the underlying sequential latent variable models (Section \ref{sec:related_work}).
    \item Our solution achieves state of the art rate-distortion performance when compared with other learned video compression approaches and performs competitively with or outperforms AVC (H.264) and HEVC (H.265) traditional codecs under equivalent settings (Section \ref{sec:experiments}).
\end{enumerate}

\section{Methodology}\label{sec:methodology}
\subsection{Problem setup}\label{sec:problem_setup}
Let us denote the image frames of a video as $\v{x}=\{\v{x}_i\}_{i\in \mathbb{N}}$. Compression of the video is done by an autoencoder that maps $\v{x}$, through an encoder $f_\text{enc}$, into compact discrete latent codes $\v{z}=\{\v{z}_i\}_{i\in \mathbb{N}}$. The codes are then used by a decoder $f_\text{dec}$ to form reconstructions $\widehat{\v{x}}=\{\widehat{\v{x}}_i\}_{i\in\mathbb{N}}$. 

We assume the use of an entropy coder together with a probabilistic model on $\v{z}$, denoted as $\mathbb{P}_\v{Z}(\cdot)$, to losslessly compress the discrete latents. The ideal codeword length can then be characterized as $R(\v{z})=-\log\mathbb{P}_{\v{Z}}(\v{z})$
which we refer to as the rate term\footnote{For practical entropy coder, there is a constant overhead per block/stream, which is negligible with a large number of bits per stream and thus can be ignored. For example, for adaptive arithmetic coding (AAC), there is up to 2-bit inefficiency per stream \cite{source_coding}.}.

Given a distortion metric $D:\v{X}\times\v{X}\to\mathbb{R}$, the lossy compression problem can be formulated as the optimization of the following Lagrangian functional

\begin{align}
\label{eq:loss_RD}
    \min_{
    f_\text{enc},
    f_\text{dec},\mathbb{P}_\v{Z}
    } \mathcal{L}_\text{RD}
    \triangleq \min_{
    f_\text{enc},
    f_\text{dec},\mathbb{P}_\v{Z}
    }\sum_{\v{x}} D(\v{x}, \widehat{\v{x}}) + \beta R(\v{z}),
\end{align}
where $\beta$ is the Lagrange multiplier that controls the balance of rate and distortion. 
It is known that this objective function is equivalent to the evidence lower bound in $\beta$-VAE\cite{betavae2017iclr} when the encoder distribution is deterministic or has a fixed entropy. Hence $\mathbb{P}_\v{Z}$ is often called the \emph{prior} distribution. We refer the reader to \cite{Habibian_2019_ICCV,stephan_2019_neurips,theisLossyImageCompression2017} for more detailed discussion. Throughout this work we use MS-SSIM \cite{wangMultiScaleStructuralSimilarity2003} measured in RGB space as our distortion metric for both training and evaluation. 

\subsection{Overview of the proposed method} \label{sec:overview_of_the_proposed_method}
In our work, we focus on the problem of \emph{online} compression of video using a \emph{causal} autoencoder, \ie one that outputs codes and a reconstruction for each frame on the fly without the need to access future context. In classic video codec terms, we are interested in the \emph{low delay P (LDP)} setting where only I-frames\footnote{We refer the reader to \cite{digital_video_introduction}, \cite{sullivanOverviewHighEfficiency2012} and Section 2 of \cite{wuVideoCompressionImage2018} for a good overview of frame structures in classic codecs.
} (Intra-coded; independent of other frames) and P-frames (Predictive inter-coded; using previously reconstructed past but not future frames) are used. 
We do not make use of B-frames (Bidirectionally interpolated frames). 

The full video sequence is broken down into \emph{groups of pictures} (GoP) $\v{x}=\{\v{x}_0, \v{x}_1, \ldots, \v{x}_{N-1}\}$ that starts with an I-frame and is followed by $N-1$ P-frames. 
We use a separate encoder, decoder and prior for the I-frames and P-frames.
The rate term is then decomposed as
\begin{align}
    R(\v{z}) = -\log \mathbb{P}_{\v{Z}}^\text{I}(\v{z}_0) - \log \mathbb{P}_{\v{Z}}^\text{P}(\v{z}_1, \ldots, \v{z}_{N-1}|\v{z}_0)
    \label{eq:rate},
\end{align}
where a superscript is used to indicate frame type.

\subsection{Network architecture}

I-frame compression is equivalent to image compression and there are already many schemes available \cite{balle2016end,balleVARIATIONALIMAGECOMPRESSION2018,mentzerConditionalProbabilityModels2018,rippel2017real,theisLossyImageCompression2017}. Here we applied the encoder and decoder design of \cite{mentzerConditionalProbabilityModels2018} for I-frame compression. Subsequently, we only focus on the design of the P-frame compression network and discuss how previously transmitted information can be best utilized to reduce the amount of information needed to reconstruct subsequent frames.

\subsubsection{Baseline architecture}\text{ }

\begin{figure}[t]
 \centering
 \includegraphics[scale=0.9]{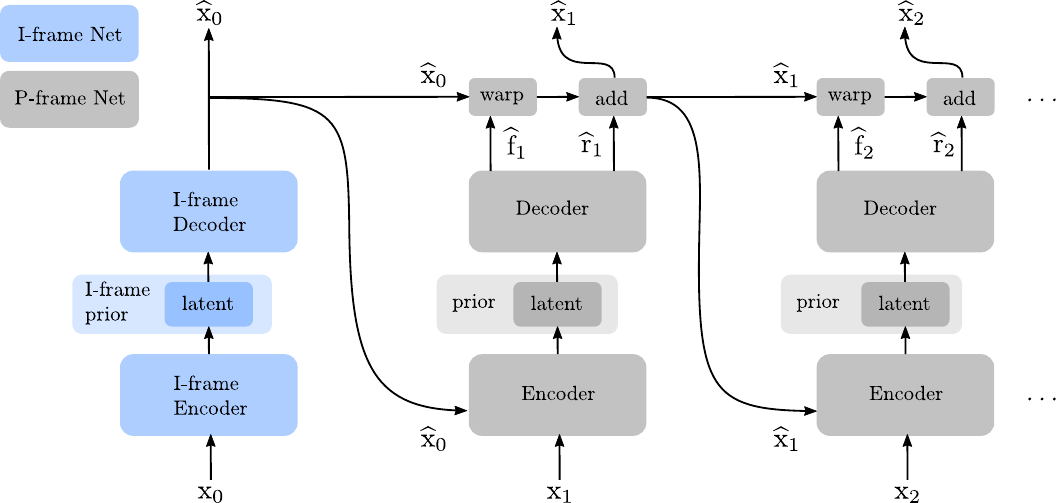}
 \caption{Base architecture of our video compression network. I-frame $\v{x}_0$ is compressed using a stand-alone image compression autoencoder. The subsequent P-frames are compressed with previous reconstructed frame as input and decoded with optical flow based motion compensation plus residual.}
  
 \label{fig:base_architecture}
\end{figure}

One straightforward way to utilize history information
is
for the autoencoder to transmit only
a motion vector (\eg block based or a dense optical field) and a residual while using the previously decoded frame(s) as a reference. 
At the decoder side, decoded motion vector (in our case optical flow) is used to warp previously decoded frame(s) using bilinear interpolation \cite{jaderberg2015spatial} (referred to later as \texttt{warp} function), which is then refined by the decoded residual. This framework serves as the basic building block for many conventional codecs such as AVC and HEVC. One instantiation of such framework built with autoencoders 
is illustrated in Fig.~\ref{fig:base_architecture}. 
Here an autoencoder based image compression network, termed I-frame network, is used to compress and reconstruct the I-frame $\v{x}_0$ independent of any other frames. Subsequent P-frames $\v{x}_t$ are processed with a separate autoencoder, termed P-frame network, that takes both the current input frame and the previous reconstructed frame as encoder inputs and produces the optical flow tensor $\widehat{\v{f}}_t$ and the residual $\widehat{\v{r}}_t$ as decoder output. 
The architecture of our P-frame network's encoder and decoder is the same as the I-frame. 
Two separate prior models are used for the entropy coding of the discrete latents in I-frame and P-frame networks. The frame is eventually reconstructed as $\widehat{\v{x}}_t\triangleq \warp(\widehat{\v{f}}_t,\widehat{\v{x}}_{t-1})+\widehat{\v{r}}_t$. 

For the P-frame network, the latent at each time $t$ contains information about $(\widehat{\v{f}}_t,\widehat{\v{r}}_t)$. One source of inefficiency, then, is that the current time step's $(\widehat{\v{f}}_t,\widehat{\v{r}}_t)$ may still exhibit temporal correlation with the past values $(\widehat{\v{f}}_{<t},\widehat{\v{r}}_{<t})$, which is not exploited. This issue remains even if we consider multiple frames as references.

To explicitly equip the network with the capability to utilize the redundancy with respect to $(\widehat{\v{f}}_{t-1},\widehat{\v{r}}_{t-1})$,
we would need to expose $(\widehat{\v{f}}_{t-1},\widehat{\v{r}}_{t-1})$ as another input, besides $\v{x}_{t-1}$, to both the encoder and the decoder for the operation at time step $t$. 
In this case the latents would only need to carry information regarding the
incremental difference between the flow field and residual between two consecutive steps, i.e. $(\widehat{\v{f}}_{t-1},\widehat{\v{r}}_{t-1})$ and $(\widehat{\v{f}}_{t},\widehat{\v{r}}_{t})$, which would lead to a higher degree of compression. We could follow the same principle to utilize even higher order redundancies but it inevitably leads to a more complicated architecture design.

\subsubsection{Feedback recurrent module}\text{ }

\begin{figure}[t]
    \centering
    \subfloat[]{{\includegraphics[scale=1]{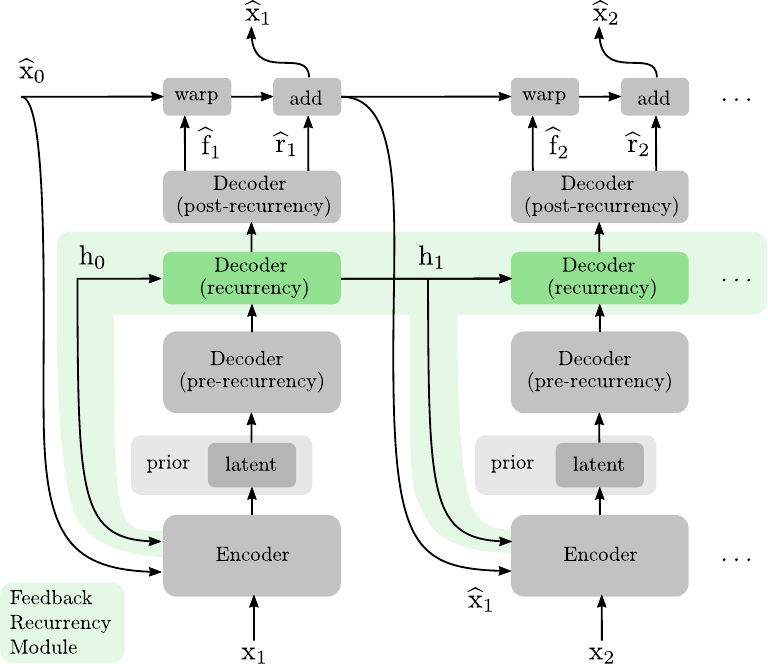}}}
    \qquad
    \subfloat[]{{\includegraphics[scale=1]{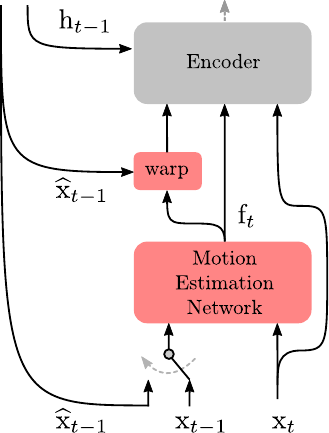}}}
    \caption{(a)Video compression network with feedback recurrent module. Detailed network architecture for the encoder and decoder is described in Appendix~\ref{appendix:model_and_training_details} 
    (b) Encoder equipped with $\menet$, an explicit optical flow estimation module.
    The switch between $\widehat{\v{x}}_{t-1}$ and $\v{x}_{t-1}$ is described at the end of Section~\ref{sec:methodology}.}
    \label{fig:complete_network}
\end{figure}

As we are trying to utilize higher order redundancy, we need to provide both the encoder and the decoder with a more suitable decoder history context. 
This observation motivates 
the use of a \emph{recurrent} neural network that is meant to accumulate and summarize relevant information received previously by the decoder, and a decoder-to-encoder \emph{feedback} connection that makes the recurrent state available at the encoder \cite{FRAE} -- see Fig.~\ref{fig:complete_network}(a). We refer to the added component as the \emph{feedback recurrent module}.

This module can be viewed as a \emph{non-linear predictive coding} scheme, where the decoder predicts the next frame based on the current latent as well as a summary of past latents.
Because the recurrent state is available to both encoder and decoder, the encoder is aware of the state of knowledge of the decoder, enabling it to send only complementary information.

The concept of feedback recurrent connection is present in several existing works: In \cite{FRAE}, the authors refer to an autoencoder with the feedback recurrent connection as FRAE and demonstrate its effectiveness in speech compression in comparison to different recurrent architectures. In \cite{Gregor2015,Gregor2016}, the focus is on progressive coding of images, and the decoder to encoder feedback is adopted for an iterative refinement of the image canvas. In the domain of video compression, \cite{rippelLearnedVideoCompression2018} proposes the concept of \emph{state-propagation}, where the network maintains a state tensor $S_t$ that is available at both the encoder and the decoder and updated by summing it with the decoder output in each time step (Fig.~5 in \cite{rippelLearnedVideoCompression2018}). In our network architecture, we propose the use of generic convolutional recurrent modules such as Conv-GRU \cite{convlstm} for the modeling of history information that is relevant for the prediction of the next frame. In Appendix~\ref{appendix:theoretical_justification}, we show the necessity of such feedback connection from an information theoretical point of view.

\subsubsection{Latent quantization and prior model}\text{ }

Given that the encoder needs to output discrete values, 
the quantization method applied needs to allow gradient based optimization.
Two popular approaches are: (1) Define a learnable codebook, and use nearest neighbor quantization in the forward pass and a differentiable softmax in the backward pass \cite{mentzerConditionalProbabilityModels2018,Habibian_2019_ICCV}, or (2) Add uniform noise to the continuous valued encoder output during training and use hard quantization at evaluation \cite{balle2016end,balleVARIATIONALIMAGECOMPRESSION2018,stephan_2019_neurips}. For the second approach, the prior distribution is often characterized by a learnable monotonic CDF function $f$, where the probability mass of a single latent value $z$ is evaluated as $f(z+1/2)-f(z-1/2)$. In this work we adopt the first approach.

As illustrated in Fig.~\ref{fig:complete_network}(a), a time-independent prior model is used. In other words, there is no conditioning of the prior model on the latents from previous time steps, and $\mathbb{P}_{\v{Z}}^\text{P}(\v{z}_1, \ldots, \v{z}_{N-1}|\v{z}_0)$ in Eq.~\eqref{eq:rate} is factorized as $\prod_{i=1}^{N-1}\mathbb{P}^P_{\v{Z}}(\v{z}_i)$. In Section~\ref{sec:related_work} we show that such factorization does not limit the capability of our model to capture any empirical data distribution.

A gated PixelCNN \cite{vandenOord2016pixelCNN} is used to model $\mathbb{P}_\v{Z}^P$, with the detailed structure of the model built based on the description in Fig.~10 of \cite{Habibian_2019_ICCV}. The prior model for the I-frame network $\mathbb{P}_\v{Z}^I$ is a separate network with the same architecture.

\subsubsection{Explicit optical flow estimation module}\text{ }
\label{sec:explicit-optical-flow-estimation-module}

In the architecture shown in Fig.~\ref{fig:complete_network}(a), the optical flow estimate tensor $\widehat{\v{f}}_t$
is produced explicitly at the end of the decoder.
When trained with the loss function $\mathcal{L}_\text{RD}$ in Eq.~\eqref{eq:loss_RD}, the learning of the optical flow estimation is only incentivized implicitly via how much that mechanism helps with the rate-effective reconstruction of the original frames.
In this setting we empirically found the decoder was almost solely relying on the residuals $\widehat{\v{r}}_t$ to form the frame reconstructions.
This observation is consistent with the observations of other works \cite{luDVCEndtoendDeep2018,rippelLearnedVideoCompression2018}. 
This problem is usually addressed by using pre-training or more explicit supervision for the optical flow estimation task.
We encourage reliance on both optical flow and residuals and facilitate optical flow estimation by (i) equipping the encoder with a U-Net~\cite{UNet} sub-network called \emph{Motion Estimation Network} ($\menet$), and (ii) introducing additional loss terms to explicitly encourage optical flow estimation.

$\menet$ estimates the optical flow $\v{f}_t$ between the current input frame $\v{x}_t$ and the previously reconstructed frame $\prevrecon$. Without $\menet$, the encoder is provided with $\curframe$ and $\prevrecon$, and is supposed to estimate the optical flow and the residuals and encode them, all in a single network. However, when attached to the encoder, $\menet$ provides the encoder directly with the estimated flow $\v{f}_t$ and the previous reconstruction warped by the estimated flow  $\warp(\prevrecon, \v{f}_t)$. In this scenario, all the encoder capacity is dedicated to the encoding task.
A schematic view of this explicit architecture is shown in Fig.~\ref{fig:complete_network}(b) and the implementation details are available in Appendix~\ref{appendix:model_and_training_details}.

When $\menet$ is integrated with the architecture in Fig.~\ref{fig:complete_network}(a), optical flow is originally estimated using $\menet$ denoted as $\v{f}_t$ and later reconstructed in the decoder denoted as $\widehat{\v{f}}_t$. In order to alleviate the problem with optical flow learning using rate-distortion loss only, we incentivize the learning of $\v{f}_t$ and $\widehat{\v{f}}_t$ via two additional dedicated loss terms,
\begin{align}
    \mathcal{L}_\text{fe}=D(\warp(\prevrecon, \v{f}_t), \curframe), 
    \mathcal{L}_\text{fd}=D(\warp(\prevrecon, \widehat{\v{f}}_t), \curframe). \notag
\end{align}
Hence the total loss we start the training with is $\mathcal{L} = \mathcal{L}_\text{RD} + \mathcal{L}_\text{fe} + \mathcal{L}_\text{fd}$ where $\mathcal{L}_\text{RD}$ is the loss function as per Eq.~\eqref{eq:loss_RD}.
We found that it is sufficient if we apply the losses $\mathcal{L}_\text{fe}$ and $\mathcal{L}_\text{fd}$ 
only at the beginning of training,
for the first few thousand iterations,
and then we revert to using just $\mathcal{L} = \mathcal{L}_\text{RD}$ and let the estimated flow be adapted to the main task, which has been shown to improve the results across variety of tasks utilizing optical flow estimation \cite{Xue_2019}.

The loss terms $\mathcal{L}_\text{fe}$ and $\mathcal{L}_\text{fd}$ are defined as the distortion between $\curframe$ and the warped version of $\prevrecon$. 
However, early in the training, $\prevrecon$ is inaccurate and as a result, such distortion is not a good choice of a learning signal for the $\menet$ or the autoencoder. 
To alleviate this problem, early in the training we use $\prevframe$ instead of $\prevrecon$ in both $\mathcal{L}_\text{fe}$ and $\mathcal{L}_\text{fd}$. 
This transition is depicted in Fig.~\ref{fig:complete_network}(b) with a switch.
It is worth to mention that the tensor fed into the encoder is always a warped previous reconstruction $\warp(\prevrecon, \v{f}_t)$, 
never a warped previous ground truth frame $\warp(\v{x}_{t-1}, \v{f}_t)$.

\section{Graphical Model Analysis} 
\label{sec:related_work}

\begin{figure}[hbt!]
 \centering
 \includegraphics[width=1\linewidth]{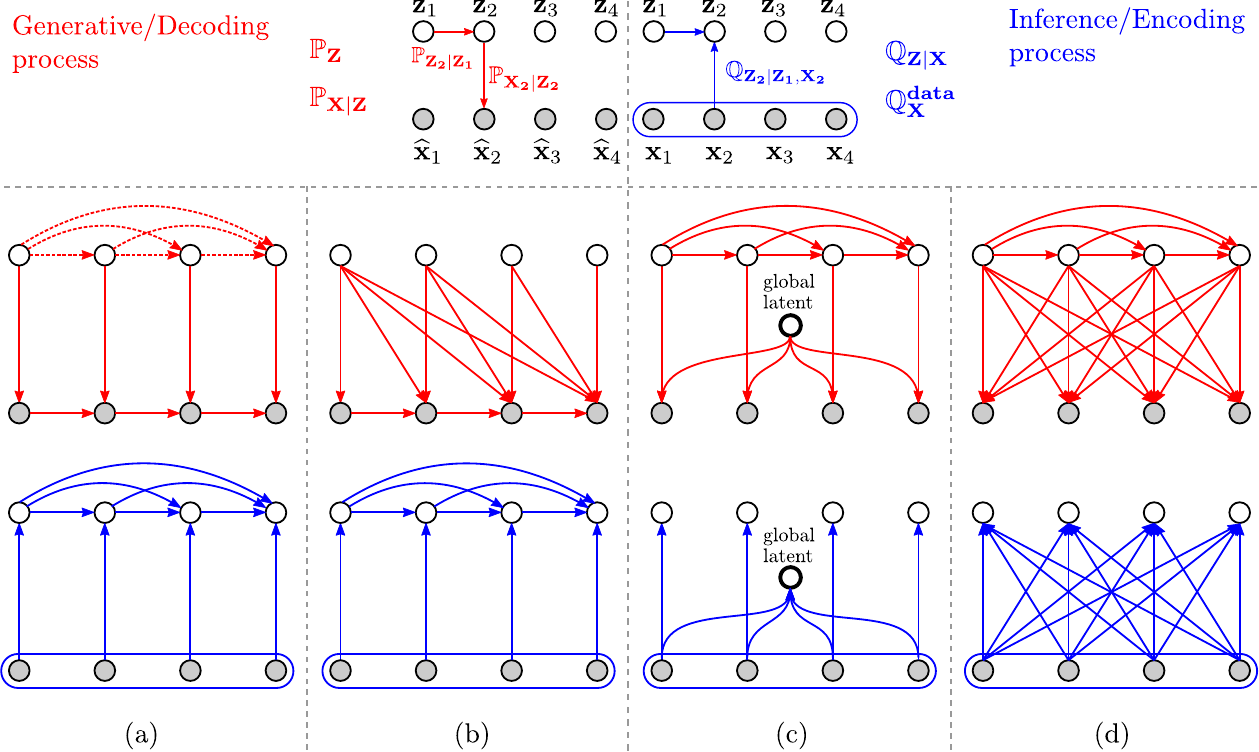}
 \caption{
 Different combinations of sequential latent variable models and inference models proposed in literature. 
 (a) without dashed lines is adopted in Lu \etal \cite{luDVCEndtoendDeep2018}. 
 (a) with dashed lines is adopted in Liu \etal \cite{liu2020learned}. 
 (b) describes both Rippel \etal \cite{rippelLearnedVideoCompression2018} and our approach. 
 (c) is used in Han \etal \cite{stephan_2019_neurips}. 
 (d) is used in Habibian \etal \cite{Habibian_2019_ICCV}.
 The blue box around all $\v{x}_{1:4}$ nodes means that they all come from a single joint distribution $\mathbb{Q}_{\v{X}}^{\operatorname{data}}\left(\v{x}_{1:4}\right)$ and that we make no assumptions on the conditional structure of that distribution. See Appendix~\ref{appendix:graphical_modelling_considerations} for detailed reasoning about these graphical models.
 }
   
 \label{fig:graphic_model_related_work}
\end{figure}
Recent success of autoencoder based learned image compression \cite{balle2016end,balleVARIATIONALIMAGECOMPRESSION2018,mentzerConditionalProbabilityModels2018,rippel2017real,theisLossyImageCompression2017} (see \cite{hu2020learning} for an overview) 
has demonstrated
neural networks' capability in modeling spatial correlations from data and the effectiveness of end-to-end joint rate-distortion training. It has motivated the use of autoencoder based solution to further capture the temporal correlation in the application of video compression and there has since been many designs for learned video compression algorithms
\cite{FRAE,rippelLearnedVideoCompression2018,luDVCEndtoendDeep2018,liu2020learned,Habibian_2019_ICCV,stephan_2019_neurips}. 
These designs differ in many detailed aspects: the technique used for latent quantization; the specific instantiation of encoder and decoder architecture; the use of atypical, often domain-specific operations beyond convolution such as Generalized Divisive Normalization (GDN) \cite{balleDensity2015,balle2016end} or Non-Local Attention Module (NLAM) \cite{liuNLAICImageCompression2019}. 
Here we leave out the comparison of those aspects and focus on one key design element: the graphical model structure of the underlying sequential latent variable model \cite{koller2009probabilistic}.

As we have briefly discussed in Section~\ref{sec:problem_setup}, the rate-distortion training objective has strong connections to amortized variational inference. In the special case of $\beta=1$, the optimization problem is equivalent to the minimization of 
    $D_\text{KL}\left(\mathbb{Q}_{\text{X}}^\text{data}\mathbb{Q}_{\text{Z}|\text{X}}|| \mathbb{P}_{\text{Z}}\mathbb{P}_{\text{X}|\text{Z}} \right)\notag$
\cite{broken_elbo} where 
$\mathbb{P}_{\text{Z}}\mathbb{P}_{\text{X}|\text{Z}}$ describes a sequential generative process  and $\mathbb{Q}_{\text{X}}^\text{data}\mathbb{Q}_{\text{Z}|\text{X}}$ can be viewed as a sequential inference process. 
$\mathbb{P}_{\text{X}|\text{Z}}$ denotes the decoder distribution induced by our distortion metric (see section 4.2 of \cite{Habibian_2019_ICCV}), 
$\mathbb{Q}_{\text{X}}^\text{data}$ denotes the empirical distribution of the training data, 
and 
$\mathbb{Q}_{\text{Z}|\text{X}}$ denotes the encoder distribution induced on the latents, in our case a Dirac-$\delta$ distribution because of the deterministic mapping from the encoder inputs to the value of the latents.

To favor the minimization of the KL divergence, we want to design $\mathbb{P}_{\text{Z}}$, $\mathbb{P}_{\text{X}|\text{Z}}$, $\mathbb{Q}_{\text{Z}|\text{X}}$ to be flexible enough so that (i) the marginalized data distribution $\mathbb{P}_{\text{X}}\triangleq\sum_{\text{Z}}\mathbb{P}_{\text{Z}}\mathbb{P}_{\text{X}|\text{Z}}$ is able to capture complex temporal dependencies in the data, and (ii) the inference process $\mathbb{Q}_{\text{Z}|\text{X}}$ encodes all the dependencies embedded in $\mathbb{P}_{\text{Z}}\mathbb{P}_{\text{X}|\text{Z}}$ \cite{webb2019inversion}. 
In Fig.~\ref{fig:graphic_model_related_work} we highlight four combinations of sequential latent variable models and the inference models proposed in literature.

Fig.~\ref{fig:graphic_model_related_work}(a) without the dashed line describes the scheme proposed in Lu \etal \cite{luDVCEndtoendDeep2018} (and the low latency mode of HEVC/AVC) where prior model is fully factorized over time and each decoded frame is a function of the previous decoded frame $\widehat{\v{x}}_{t-1}$ and the current latent code $\v{z}_t$. 
There are two major limitations of this approach: firstly, the marginalized sequential distribution $\mathbb{P}_{\text{X}}$ is confined to follow Markov property $\mathbb{P}_\text{X}(\v{x}_{1:T})=\mathbb{P}_{\text{X}_1}(\v{x}_1)\prod_{t=2}^T \mathbb{P}_{\text{X}_t|\text{X}_{t-1}}(\v{x}_t|\v{x}_{t-1})$;
secondly, it assumes that the differences in subsequent frames (\eg optical flow and residual) are independent across time steps. 
To overcome these two limitations, Liu \etal \cite{liu2020learned} propose an auto-regressive prior model through the use of ConvLSTM over time, which corresponds to Fig.~\ref{fig:graphic_model_related_work}(a) including dashed lines. 
In this case, the marginal $\mathbb{P}_\text{X}(\v{x}_{1:T})$ is fully flexible in the sense that it does not make any conditional independence assumptions between $\v{x}_{1:T}$, 
see Appendix~\ref{appendix:full_flexibility_marginal} for details.

Fig.~\ref{fig:graphic_model_related_work}(b) describes another way to construct a fully flexible marginalized sequential data distribution, by having the decoded frame depend on all previously transmitted latent codes. 
Both Rippel \etal \cite{rippelLearnedVideoCompression2018} and our approach fall under this category by introducing a recurrent state in the decoder which summarizes information from the previously transmitted latent codes and processed features. 
Rippel \etal \cite{rippelLearnedVideoCompression2018} use a multi-scale state with a complex state propagation mechanism, 
whereas we use a convolutional recurrency module (ConvGRU) \cite{convlstm}.
In this case, $\mathbb{P}_\text{X}(\v{x}_{1:T})$ is also fully flexible, 
details in Appendix~\ref{appendix:full_flexibility_marginal}.

The latent variable models in Fig.~\ref{fig:graphic_model_related_work}(c) and (d) break the causality constraint,
\ie the encoding of a GoP can only start when all of its frames are available.
In (c), a global latent is used to capture time-invariant features, which is adopted in Han \etal \cite{stephan_2019_neurips}. 
In (d), the graphical model is fully connected between each frame and latent across different time steps, which described the scheme applied in Habibian \etal \cite{Habibian_2019_ICCV}.

One advantage of our approach (as well as Rippel \etal \cite{rippelLearnedVideoCompression2018}), which corresponds to Fig.~\ref{fig:graphic_model_related_work}(b), is that the prior model can be fully factorized across time, $\mathbb{P}_{\text{Z}}(\v{z}_{1:T}) = \prod_{t=1}^T \mathbb{P}_{\text{Z}}(\v{z}_t)$,
without compromising the flexibility of the marginal distribution $\mathbb{P}_\text{X}(\v{x}_{1:T})$. 
Factorized prior allows \emph{parallel} entropy decoding of the latent codes across time steps (but potentially still \emph{sequential} across dimensions of $\v{z}_t$ within each time step). 
On the inference side, in Appendix~\ref{appendix:theoretical_justification} we show that any connection from $\v{x}_{<t}$ to $\v{z}_t$, which could be modeled by adding a recurrent component on the encoder side, is not necessary.

\section{Experiments}\label{sec:experiments}

\subsection{Training setup}

\subsubsection*{Datasets}

Our training dataset was based on Kinetics400 \cite{Kinetics400}. 
We selected a subset of the high-quality videos in Kinetics (width and height over 720 pixels), took the first 16 frames of each video, and downscaled them (to remove compression artifacts) such that the smaller of the dimensions would be 256 pixels. 
At train time, random crops of size $160\times160$ were used. For validation, we used videos from HDgreetings, NTIA/ITS and SVT.

We used UVG \cite{UVG} as our test dataset. UVG contains 7 $1080$p video sequences and a total of 3900 frames. 
The raw UVG $1080$p videos are available in YUV420-8bit format
and we converted them to RGB using OpenCV \cite{opencv} for evaluation.

\subsubsection*{Training}
\label{sec:training_details}
All implementations were done in the PyTorch framework \cite{PyTorch}. 
The distortion measure $1-$MS-SSIM \cite{wangMultiScaleStructuralSimilarity2003} was normalized per frame and the rate term was normalized per pixel (i.e., bits per pixel) before they were combined $D+\beta R$. We trained 
our network 
with five $\beta$ values
$\beta=\{0.025, 0.05, 0.1, 0.2, 0.3\}$ 
to obtain a rate-distortion curve. The I-frame network and P-frame network were trained jointly from scratch without pretraining for any part of the network.

The GoP size of $8$ was used during training,
\ie the recurrent P-frame network was unrolled 7 times. 
We used a batch size of 16, for a total of 250k gradient updates (iterations) corresponding to 
20 epochs. When training with the flow enhancement network $\menet$, we applied $\mathcal{L}_\text{fe}$ and $\mathcal{L}_\text{fd}$ until 20k iterations, and the transition of the input from $\prevframe$ to $\prevrecon$ was done at iteration 15k.
All models were trained using Adam optimizer \cite{kingmaAdamMethodStochastic2015} 
with the initial learning rate of $10^{-4}$, $\beta_1=0.9$ and $\beta_2=0.999$. 
The learning rate was annealed by a factor of 0.8 every 100k iterations. Performance on the validation set was evaluated every epoch, results presented below are the model checkpoints performing best on the validation set.
See Appendix~\ref{appendix:model_and_training_details} for details on the architecture and training.

\subsection{Comparison with other methods}

In this section, we compare the performance of our method with existing learned approaches as well as popular classic codecs.

\subsubsection*{Comparison with learning-based methods} 

\begin{figure}
    \centering
    \subfloat[]{{\includegraphics[width=6cm]{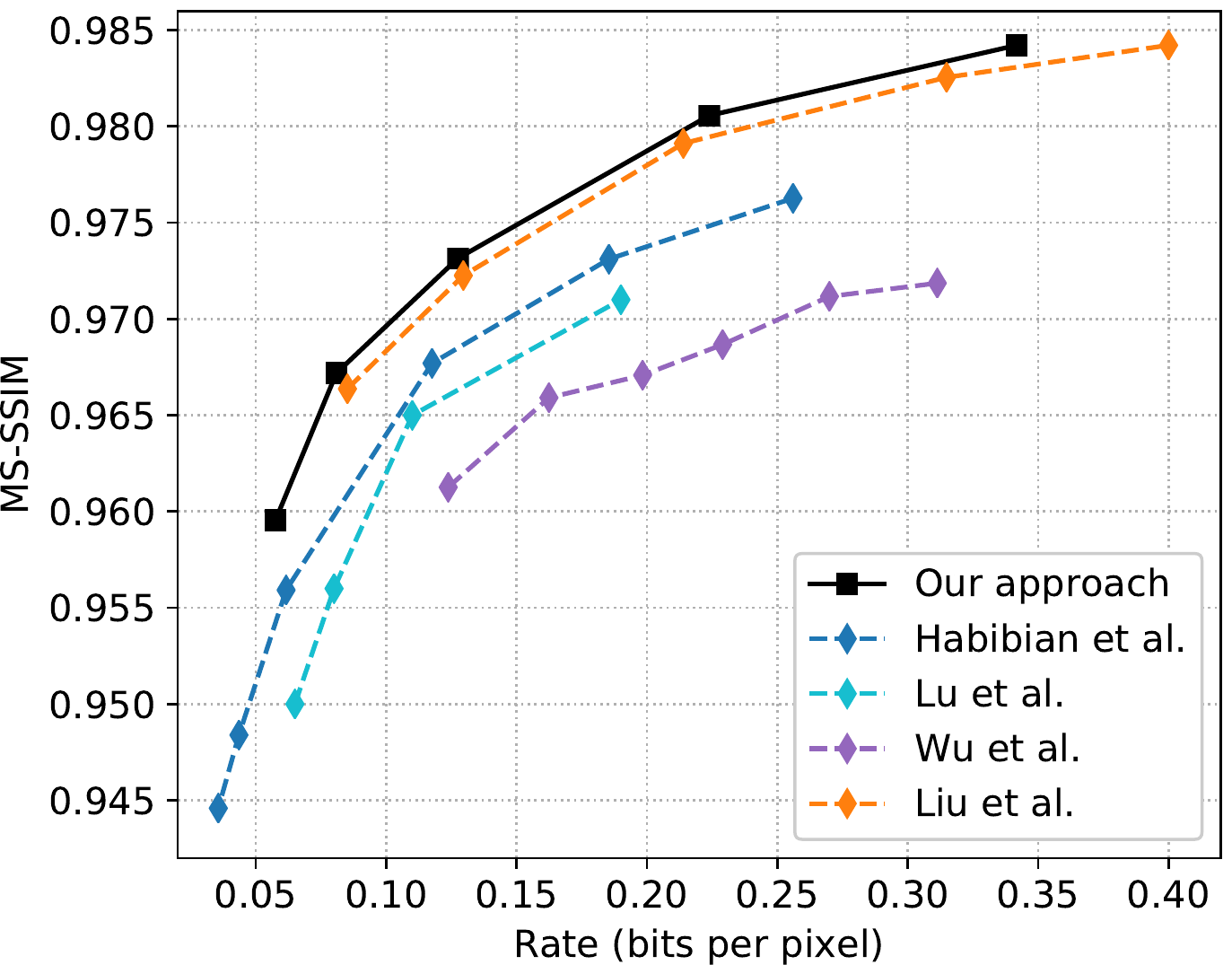}}}%
    \subfloat[]{{\includegraphics[width=6cm]{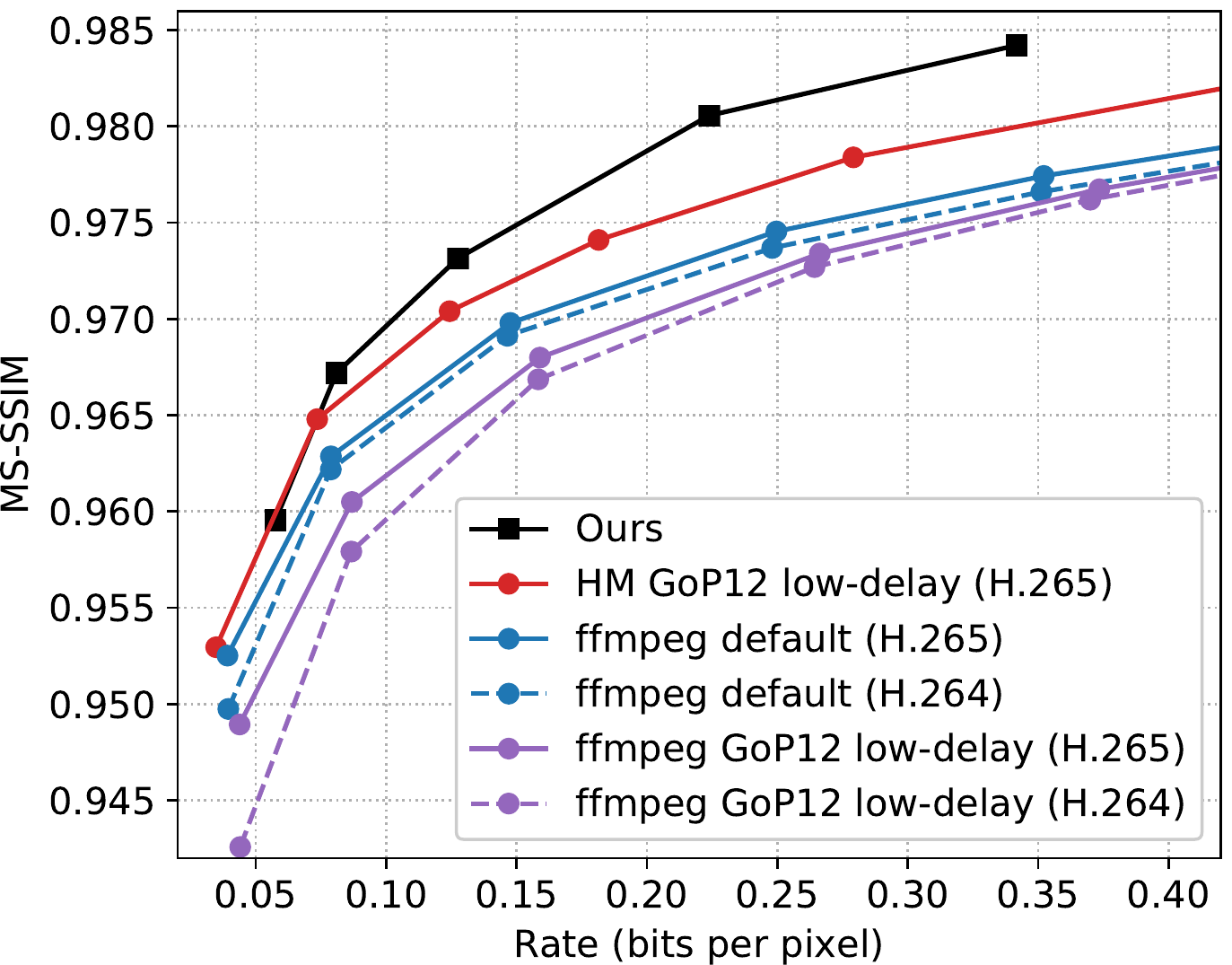}}}%
    \caption{(a) Comparison to the state-of-the-art learned methods. (b) Comparison with classic codecs. Both comparisons are on UVG $1080$p dataset.}
    \label{fig:compare_with_learned_and_classic}%
\end{figure}

In Fig.~\ref{fig:compare_with_learned_and_classic}(a), we compare the performance of our solution with several learned approaches on UVG $1080$p dataset in terms of MS-SSIM versus bitrate where MS-SSIM was first averaged per video and then averaged across different videos. Our method outperforms all the compared ones in MS-SSIM across the range of bitrate between around $0.05$bpp to around $0.35$bpp. Lu \etal \cite{luDVCEndtoendDeep2018} and Liu \etal \cite{liu2020learned} are both causal solutions and they are evaluated with GoP size of 12. Habibian \etal and Wu \etal are non-causal solutions and their results are reported based on GoP sizes of 8 and 12, respectively. Our results are evaluated at GoP of 8, same as in training.

\subsubsection*{Comparison with traditional video codecs}
We compared our method with the most popular standard codecs \ie H.265 \cite{sullivanOverviewHighEfficiency2012} and H.264~\cite{avc} on UVG $1080$p dataset. 
The results are generated with three different sets of settings (more details are provided in Appendix~\ref{sec:appendix_baseline_generation}):
\begin{itemize}
    \item \texttt{ffmpeg}~\cite{ffmpeg} implementation of H.264 and H.265 in low latency mode with GoP size of 12.
    \item \texttt{ffmpeg}~\cite{ffmpeg} implementation of H.264 and H.265 in default mode. 
    \item \texttt{HM}~\cite{HM} implementation of H.265 in low latency mode with GoP size of 12.
\end{itemize}

The low latency mode was enforced to H.264 and H.265 to make the problem settings the same as our causal model. GoP size 12 on the other hand, although different from our GoP size 8, was consistent with the settings reported in other papers and provided H.264 and H.265 an advantage as they perform better with larger GoP sizes.

The comparisons are shown in Fig.~\ref{fig:compare_with_learned_and_classic}(b) in terms of MS-SSIM versus bitrate where MS-SSIM was calculated in RGB domain. As can be seen from this figure, our model outperforms the \texttt{HM} implementation of H.265 and the \texttt{ffmpeg} implementation of H.265 and H.264 in both low latency and default settings, at bitrates above $0.09\text{bpp}$. For 1080p resolution this is the range of bitrates of interest for practitioners, \eg Netflix uses the range of about $0.09-0.13\text{bpp}$ for their 1080p resolution video streaming \cite{netflix-data-usage}.

\subsubsection*{Qualitative comparison}
\begin{figure}
\centering
\includegraphics[trim={0 .1cm 0 0},clip,width=.99\linewidth]{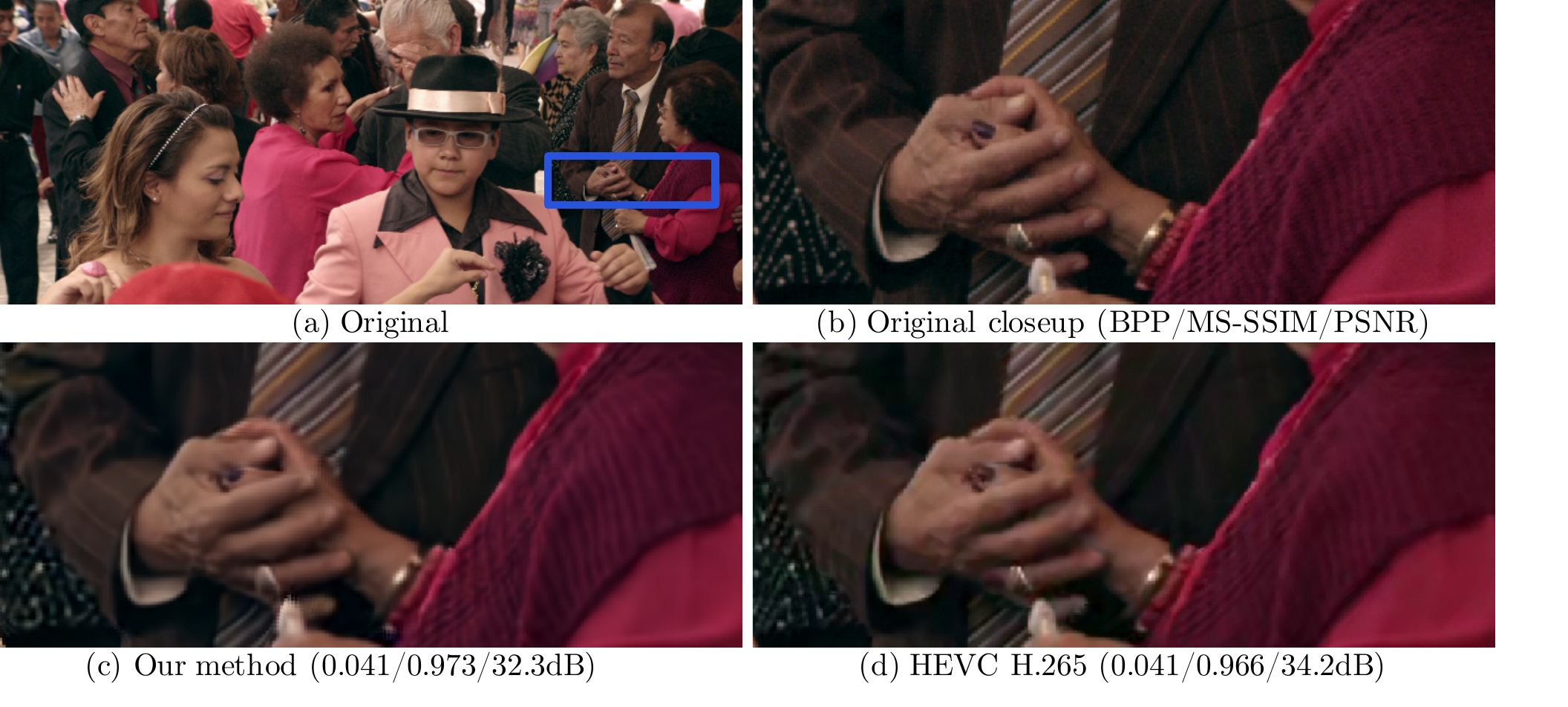}
\caption{An illustration of qualitative characteristics of our architecture versus H.265 (\texttt{ffmpeg}) at a comparable bitrate. (a) and (b) are the original frame and its closeup, (c) and (d) are the closeup on the reconstructed frames from our method and H.265. 
To make the comparison fair, we used  HEVC with fixed GoP setting (\texttt{min-keyint=8:scenecut=0}) at a similar rate, so both methods are at equal bpp and show the 4th P-frame of a GoP of 8.  Frame 229 of Tango video from Netflix Tango in Netflix El Fuente; see Appendix~\ref{appendix:image_attribution} for license information.
}
\label{fig:closeup}
\end{figure}

In Fig.~\ref{fig:closeup} we compare the visual quality of our lowest rate model with \texttt{ffmpeg} implementation of H.265 at in low latency mode a similar rate. We can see that our result is free from the blocking artifact usually present in H.265 at low bitrate -- see around the edge of fingers -- and preserves more detailed texture -- see structures of hand veins and strips on the coat. 
In Appendix~\ref{appendix:qualitative_examples}, we provide more examples with detailed error and bitrate maps.

\begin{figure}
\centering
\includegraphics[width=.99\linewidth]{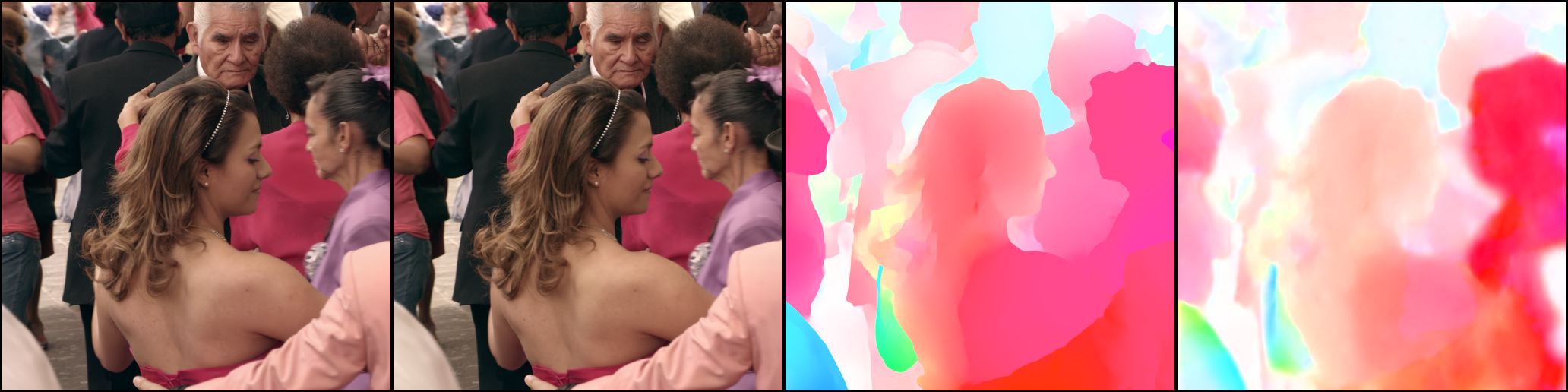}
\caption{An example of the estimated optical flow, left to right: two consecutive frames used for estimation, FlowNet2 results, our decoder output $\widehat{\v{f}}$. 
Note that the flow produced by our model is decoded from a compressed latent, and is trained to maximize total compression performance rather than warping accuracy.
See Appendix~\ref{appendix:image_attribution} for license information.
}
\label{fig:optical_flow}
\end{figure}

\begin{figure}
\centering
\includegraphics[height=3.7cm]{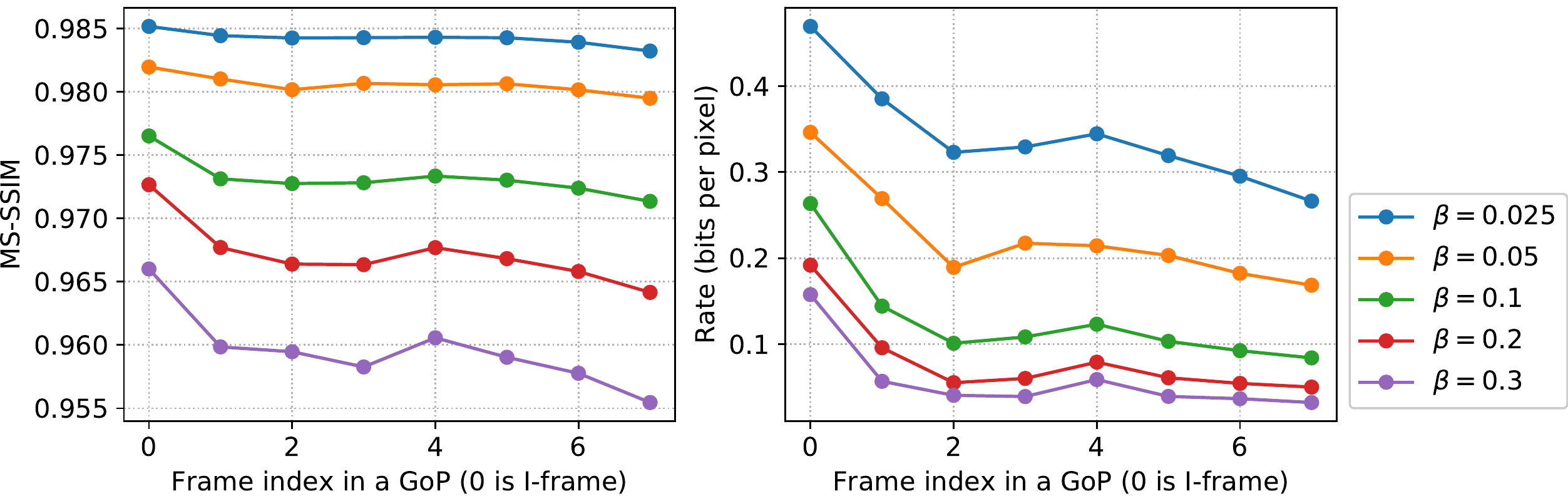}
\caption{Average MS-SSIM and rate as a function of frame-index in a GoP of 8.}
\label{fig:bpp_msssim_per_frame}
\end{figure}

As noted in Section~\ref{sec:explicit-optical-flow-estimation-module}, the dedicated optical flow enforcement loss terms were removed at a certain point and the training continued using $\mathcal{L}_\text{RD}$ only. 
As a result, our network learned a form of optical flow that contributed maximally to $\mathcal{L}_\text{RD}$ and did not necessarily follow the ground truth optical flow (if such were available). 
Fig.~\ref{fig:optical_flow} compares an instance of the optical flow learned in our network with the corresponding optical flow generated using a pre-trained FlowNet2 network~\cite{FlowNet2}.
The optical flow reconstructed by the decoder $\widehat{\v{f}}$ has larger proportion of values set to zero than the flow estimated by FlowNet2 -- this is consistent with the observations by Lu \etal \cite{luDVCEndtoendDeep2018} in their Fig. 7 where they argue that it allows the flow to be more compressible.

\subsection{Ablation study}

\begin{figure}[hb!]
\centering
\includegraphics[height=5.5cm]{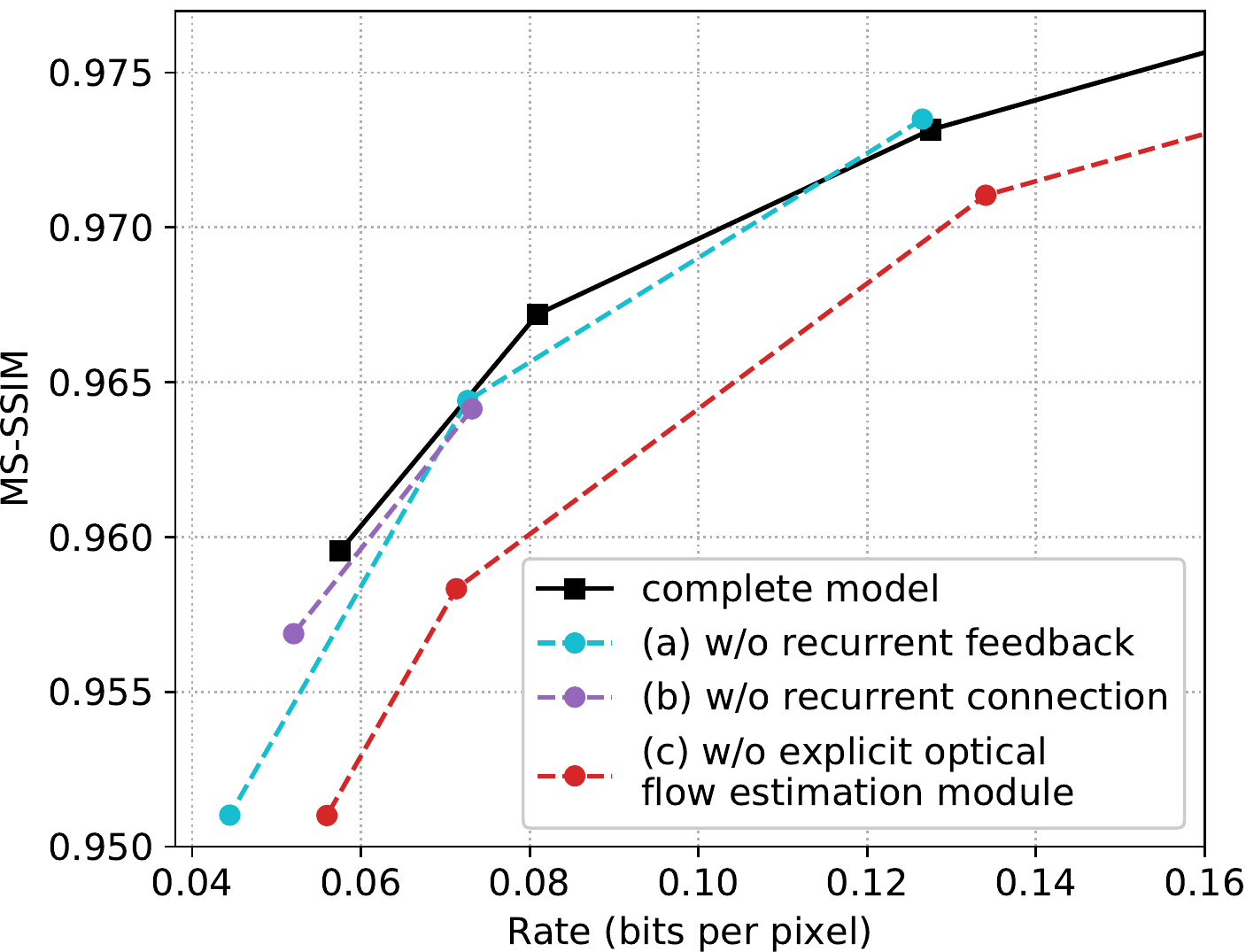}
\caption{Ablation study on UVG dataset. All models were evaluated at GoP of size 8. Model (a) is obtained by removing the decoder to encoder feedback connection in Fig.~\ref{fig:complete_network}(a). Model (b) removes the the feedback recurrent module. Model (c) removes the explicit optical flow estimation module in Fig.~\ref{fig:complete_network}(b). 
}
\label{fig:ablation}
\end{figure}

To understand the effectiveness of different components in our design, in Fig.~\ref{fig:ablation} we compare the performance after removing (a) decoder to encoder recurrent feedback, or (b) the feedback recurrent module in Fig.~\ref{fig:complete_network}(a), or (c) the explicit optical flow estimation module. 
We focus the comparison on low rate regime where the difference is more noticeable.

Empirically we find the explicit optical flow estimation component to be quite essential -- without it the optical flow output from the decoder has very small magnitude and is barely useful, resulting in consistent loss of at least 0.002 in MS-SSIM score for rate below 0.16 bpp. In comparison, the loss in performance after removing either the decoder to encoder feedback or the recurrent connection all together is minor and only show up at very low rate region below 0.1 bpp. 

Similar comparisons have been done in existing literature. Rippel \etal \cite{rippelLearnedVideoCompression2018} report large drop in performance when removing the learned state in their model (Fig.~9 of  \cite{rippelLearnedVideoCompression2018}; about 0.005 drop in MS-SSIM at bpp of 0.05), while Liu \etal \cite{liu2020learned}, which utilizes recurrent prior to exploit longer range temporal redundancies, reports relatively smaller degradation after removing their recurrent module (about 0.002 drop in MS-SSIM in Fig.~9 of \cite{liu2020learned}).

We suspect that the lack of gain we observe from the feedback recurrent module is due to both insufficient recurrent module capacity and receptive field, and plan to investigate further as one of our future studies.

\subsection{Empirical observations}

\subsubsection{Temporal consistency}\text{ }
We found, upon visual inspection, that for our low bitrate models, the decoded video displays a minor yet noticeable flickering artifact occurring at the GoP boundary, which we attribute to two factors.

Firstly, as shown in Fig.~\ref{fig:bpp_msssim_per_frame}, there is a gradual degradation of P-frame quality as it moves further away from the I-frame. 
It is also apparent from Fig.~\ref{fig:bpp_msssim_per_frame} that the rate allocation tends to decline as P-frame network unrolls, which compounds the degradation of P-frame quality.  
This shows that the end-to-end training of frame-averaged rate-distortion loss does not 
favor temporal consistency of frame quality,
which makes it difficult to apply the model with larger GoP sizes and motivates the use of temporal-consistency related losses \cite{Gao_2019_ReCoNet,Lai-ECCV-2018}.

Second part of the problem is due to the nature of the MS-SSIM metric and hence is not reflected in Fig.~\ref{fig:bpp_msssim_per_frame}. This aspect is described in the following section.

\subsubsection{Color shift}\text{ }
Another empirical observation is that the MS-SSIM metric seems to be partially invariant to uniform color shift in flat regions and due to that the low bitrate models we have trained are prone to exhibit slight color shift, similar to what has been reported in \cite{l1_plus_msssim}. This color change is sometimes noticeable during I-frame to P-frame transitioning and contributes to the flickering artifact. One item for future study is to find suitable metrics or combination of metrics that captures both texture similarity and color fidelity.
See Appendix~\ref{appendix:msssim_color_issue} for details.

\section{Conclusion}
\label{sec:conclusion}
In this paper, we proposed a new autoencoder based learned lossy video compression solution featuring a feedback recurrent module and an explicit optical flow estimation module. It achieves state of the art performance among learned video compression methods and delivers comparable or better rate-distortion when compared with classic codecs in low latency mode. In future work, we plan to improve its temporal consistency, solve the color fidelity issue, and reduce computational complexity, paving the way to a practical real-time learned video codec.

\clearpage
%
%
\bibliographystyle{splncs04}
\bibliography{sections/bib_short_strings,sections/references}

\begin{thebibliography}{10}
\providecommand{\url}[1]{\texttt{#1}}
\providecommand{\urlprefix}{URL }
\providecommand{\doi}[1]{https://doi.org/#1}

\bibitem{video-data-global-usage}
2019 global internet phenomena report.
  \url{https://www.ncta.com/whats-new/report-where-does-the-majority-of-internet-traffic-come},
  accessed: 2020-02-28

\bibitem{digital_video_introduction}
Digital video introduction.
  \url{https://github.com/leandromoreira/digital_video_introduction/blob/master/README.md#frame-types},
  accessed: 2020-03-02

\bibitem{ffmpeg}
ffmpeg. \url{http://ffmpeg.org/}, accessed: 2020-02-21

\bibitem{HM}
{High Efficiency Video Coding (HEVC)}. \url{https://hevc.hhi.fraunhofer.de/},
  accessed: 2020-02-21

\bibitem{netflix-data-usage}
How much data does {Netflix} use?
  \url{https://www.howtogeek.com/338983/how-much-data-does-netflix-use/},
  accessed: 2020-02-28

\bibitem{UVG}
{Ultra Video Group} test sequences. \url{http://ultravideo.cs.tut.fi/},
  accessed: 2020-02-21

\bibitem{broken_elbo}
Alemi, A.A., Poole, B., Fischer, I., Dillon, J.V., Saurous, R.A., Murphy, K.:
  {Fixing a Broken ELBO}. In: ICML (2018)

\bibitem{balle2016end}
Ball\'e, J., Laparra, V., Simoncelli, E.P.: {End-to-End Optimized Image
  Compression}. In: ICLR (2017)

\bibitem{balleVARIATIONALIMAGECOMPRESSION2018}
Ball\'e, J., Minnen, D., Singh, S., Hwang, S.J., Johnston, N.: {Variational
  Image Compression with a Scale Hyperprior}. In: ICLR (2018)

\bibitem{balleDensity2015}
Ballé, J., Laparra, V., Simoncelli, E.P.: {Density Modeling of Images using a
  Generalized Normalization Transformation}. In: ICLR (2016)

\bibitem{opencv}
Bradski, G.: {The OpenCV Library}. Dr. Dobb's Journal of Software Tools  (2000)

\bibitem{Gao_2019_ReCoNet}
Gao, C., Gu, D., Zhang, F., Yu, Y.: {ReCoNet: Real-Time Coherent Video Style
  Transfer Network}. In: ACCV (2019)

\bibitem{Gregor2016}
Gregor, K., Besse, F., Rezende, D.J., Danihelka, I., Wierstra, D.: {Towards
  Conceptual Compression}. In: NeurIPS (2016)

\bibitem{Gregor2015}
Gregor, K., Danihelka, I., Graves, A., Rezende, D.J., Wierstra, D.: {DRAW: A
  Recurrent Neural Network For Image Generation}. In: ICML (2015)

\bibitem{Habibian_2019_ICCV}
Habibian, A., {van Rozendaal}, T., Tomczak, J.M., Cohen, T.S.: {Video
  Compression With Rate-Distortion Autoencoders}. In: ICCV (2019)

\bibitem{stephan_2019_neurips}
Han, J., Lombardo, S., Schroers, C., Mandt, S.: {Deep Probabilistic Video
  Compression}. In: NeurIPS (2019)

\bibitem{heDeepResidualLearning2016}
He, K., Zhang, X., Ren, S., Sun, J.: {Deep Residual Learning for Image
  Recognition}. In: CVPR (2016)

\bibitem{betavae2017iclr}
Higgins, I., Matthey, L., Pal, A., Burgess, C., Glorot, X., Botvinick, M.,
  Mohamed, S., Lerchner, A.: {beta-VAE: Learning Basic Visual Concepts with a
  Constrained Variational Framework}. In: ICLR (2017)

\bibitem{hu2020learning}
Hu, Y., Yang, W., Ma, Z., Liu, J.: Learning end-to-end lossy image compression:
  A benchmark. arXiv:2002.03711  (2020)

\bibitem{FlowNet2}
Ilg, E., Mayer, N., Saikia, T., Keuper, M., Dosovitskiy, A., Brox, T.: {FlowNet
  2.0: Evolution of Optical Flow Estimation with Deep Networks}. In: CVPR
  (2017)

\bibitem{jaderberg2015spatial}
Jaderberg, M., Simonyan, K., Zisserman, A., Kavukcuoglu, K.: {Spatial
  Transformer Networks}. In: NeurIPS (2015)

\bibitem{Kinetics400}
Kay, W., Carreira, J., Simonyan, K., Zhang, B., Hillier, C., Vijayanarasimhan,
  S., Viola, F., Green, T., Back, T., Natsev, P., Suleyman, M., Zisserman, A.:
  {The Kinetics Human Action Video Dataset}. arxiv:1705.06950  (2017)

\bibitem{kingmaAdamMethodStochastic2015}
Kingma, D., Ba, J.: Adam: {{A Method}} for {{Stochastic Optimization}}. In:
  ICLR (2015)

\bibitem{koller2009probabilistic}
Koller, D., Friedman, N.: Probabilistic Graphical Models: Principles and
  Techniques. MIT Press (2009)

\bibitem{Lai-ECCV-2018}
Lai, W.S., Huang, J.B., Wang, O., Shechtman, E., Yumer, E., Yang, M.H.:
  {Learning Blind Video Temporal Consistency}. In: ECCV (2018)

\bibitem{liuNLAICImageCompression2019}
Liu, H., Chen, T., Guo, P., Shen, Q., Cao, X., Wang, Y., Ma, Z.: {Non-local
  Attention Optimized Deep Image Compression}. arXiv:1904.09757  (2019)

\bibitem{liu2020learned}
Liu, H., shen, H., Huang, L., Lu, M., Chen, T., Ma, Z.: {Learned Video
  Compression via Joint Spatial-Temporal Correlation Exploration}. In: AAAI
  (2020)

\bibitem{luDVCEndtoendDeep2018}
Lu, G., Ouyang, W., Xu, D., Zhang, X., Cai, C., Gao, Z.: {DVC: An End-to-End
  Deep Video Compression Framework}. In: CVPR (2019)

\bibitem{mentzerConditionalProbabilityModels2018}
Mentzer, F., Agustsson, E., Tschannen, M., Timofte, R., Van~Gool, L.:
  {Conditional Probability Models for Deep Image Compression}. In: CVPR (2018)

\bibitem{PyTorch}
Paszke, A., Gross, S., Massa, F., Lerer, A., Bradbury, J., Chanan, G., Killeen,
  T., Lin, Z., Gimelshein, N., Antiga, L., Desmaison, A., Kopf, A., Yang, E.,
  DeVito, Z., Raison, M., Tejani, A., Chilamkurthy, S., Steiner, B., Fang, L.,
  Bai, J., Chintala, S.: {PyTorch: An Imperative Style, High-Performance Deep
  Learning Library}. In: NeurIPS (2019)

\bibitem{source_coding}
Pearlman, W.A., Said, A.: {Digital Signal Compression: Principles and
  Practice}. Cambridge University Press (2011)

\bibitem{rippel2017real}
Rippel, O., Bourdev, L.: {Real-time adaptive image compression}. In: ICML
  (2017)

\bibitem{rippelLearnedVideoCompression2018}
Rippel, O., Nair, S., Lew, C., Branson, S., Anderson, A.G., Bourdev, L.:
  {Learned Video Compression}. In: ICCV (2019)

\bibitem{UNet}
Ronneberger, O., Fischer, P., Brox, T.: {U-Net: Convolutional Networks for
  Biomedical Image Segmentation}. In: MICCAI (2015)

\bibitem{convlstm}
Shi, X., Chen, Z., Wang, H., Yeung, D., Wong, W., Woo, W.: {Convolutional LSTM
  Network: A Machine Learning Approach for Precipitation Nowcasting}. In:
  NeurIPS (2015)

\bibitem{sullivanOverviewHighEfficiency2012}
Sullivan, G.J., Ohm, J.R., Han, W.J., Wiegand, T.: {Overview of the High
  Efficiency Video Coding (HEVC) Standard}. IEEE Trans. Circuits Syst. Video
  Technol.  \textbf{22}(12),  1649--1668 (Dec 2012)

\bibitem{theisLossyImageCompression2017}
Theis, L., Shi, W., Cunningham, A., Husz\'ar, F.: {Lossy Image Compression with
  Compressive Autoencoders}. In: ICLR (2017)

\bibitem{vandenOord2016pixelCNN}
{van den Oord}, A., Kalchbrenner, N., Espeholt, L., Kavukcuoglu, K., Vinyals,
  O., Graves, A.: {Conditional Image Generation with PixelCNN Decoders}. In:
  NeurIPS (2016)

\bibitem{Wang2009MSELoveItLeaveIt}
{Wang}, Z., {Bovik}, A.C.: {Mean Squared Error: Love it or leave it? A new look
  at Signal Fidelity Measures}. In: IEEE Signal Processing Magazine. vol.~26,
  pp. 98--117 (Jan 2009)

\bibitem{wangMultiScaleStructuralSimilarity2003}
Wang, Z., Bovik, A.C., Sheikh, H.R., Simoncelli, E.P., et~al.: {Image quality
  assessment: from error visibility to structural similarity}. IEEE Trans. on
  Image Processing  \textbf{13}(4),  600--612 (2004)

\bibitem{webb2019inversion}
Webb, S., Goli\'{n}ski, A., Zinkov, R., N, S., Rainforth, T., Teh, Y.W., Wood,
  F.: {Faithful Inversion of Generative Models for Effective Amortized
  Inference}. In: NeurIPS (2018)

\bibitem{avc}
{Wiegand}, T., {Sullivan}, G.J., {Bjontegaard}, G., {Luthra}, A.: {Overview of
  the H.264/AVC video coding standard}. IEEE Transactions on Circuits and
  Systems for Video Technology  \textbf{13}(7),  560--576 (July 2003)

\bibitem{wuVideoCompressionImage2018}
Wu, C.Y., Singhal, N., Kr\"ahenb\"uhl, P.: {Video Compression through Image
  Interpolation}. In: ECCV (2018)

\bibitem{Xue_2019}
Xue, T., Chen, B., Wu, J., Wei, D., Freeman, W.T.: {Video Enhancement with
  Task-Oriented Flow}. IJCV  \textbf{127}(8) (Feb 2019)

\bibitem{FRAE}
Yang, Y., Sautière, G., Ryu, J.J., Cohen, T.S.: {Feedback Recurrent
  AutoEncoder}. In: ICASSP (2019)

\bibitem{l1_plus_msssim}
{Zhao}, H., {Gallo}, O., {Frosio}, I., {Kautz}, J.: {Loss Functions for Image
  Restoration With Neural Networks}. In: IEEE Transactions on Computational
  Imaging (2017)

\end{thebibliography}

\clearpage
\appendix
{\Large\noindent\bf Appendix}
\section{Model and training details}
\label{appendix:model_and_training_details}

\begin{figure}[hbt!]
 \centering
 \includegraphics[width=0.9\linewidth]{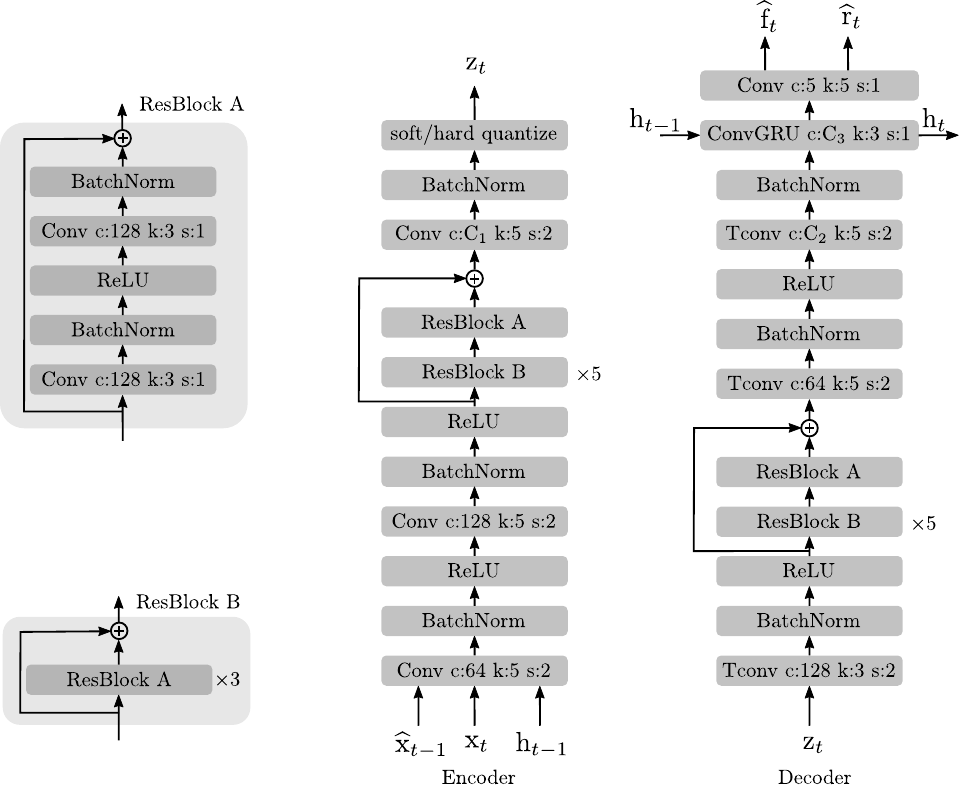}
 \caption{Architecture of our P-frame network (flow enhanced module is separately illustrated). \emph{Tconv} denotes transposed convolution. For (transposed) convolutional layer $c$ denotes the number of output channels, $k$ denotes the kernel size and $s$ denotes the stride. In all of our experiments, we set $\text{C}_1=32$, $\text{C}_2=10$, and $\text{C}_3=32$. Architecture of I-frame network differs by removing the final two layers in the decoder, setting $\text{C}_2=3$, and having $\v{x}_0$ as the only input to the encoder. Note that a large part of this architecture is inherited from Fig.~2 in \cite{mentzerConditionalProbabilityModels2018} and Fig.~9 in \cite{Habibian_2019_ICCV}.}
 \vspace{-0.5cm}
 \label{fig:detailed_architecture}
\end{figure}

\begin{figure}[hbt!]
 \centering
 \includegraphics[width=0.8\linewidth]{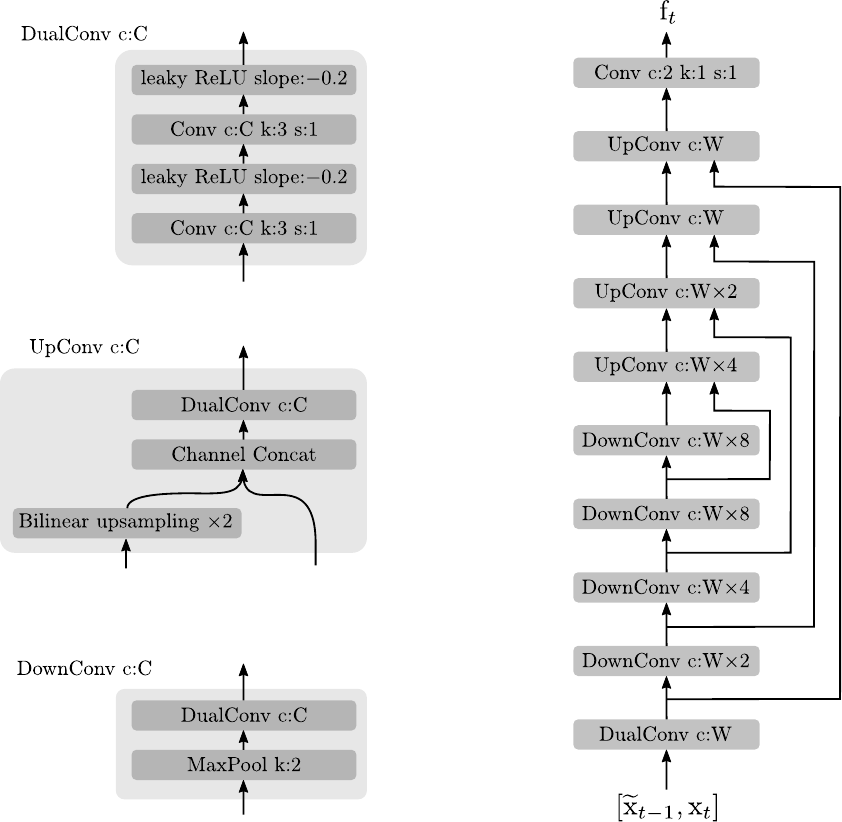}
 \caption{Architecture of our optical flow estimation network. In all of our experiments, we set $W=16$.}
 \vspace{-0.5cm}
 \label{fig:detailed_architecture_flownet}
\end{figure}

Detailed architecture of the encoder and decoder in Fig.~\ref{fig:complete_network}(a) is illustrated in Fig.~\ref{fig:detailed_architecture}. For prior modeling, we use the same network architecture as described in Section B.2 of \cite{Habibian_2019_ICCV}. Detailed architecture of the optical flow estimation network in Fig.~\ref{fig:complete_network}(b) is illustrated in Fig.~\ref{fig:detailed_architecture_flownet}. 

As can be seen from Fig.~\ref{fig:detailed_architecture}, we use BatchNorm throughout the encoder and decoder architecture to help stabilize training. However, since previous decoded frame is fed back into the encoder in the next time step, during training it is not possible to apply the same normalization across all the unrolled time steps, which will lead to an inconsistency across training and evaluation. To solve this issue, we switch BatchNorm to evaluation mode during training at iteration 40k, after which its parameters are no longer updated.

\section{Images used in figures}\label{appendix:image_attribution}

Video used in Figures~\ref{fig:optical_flow}, ~\ref{fig:closeup}, ~\ref{fig:reel_hevc}, ~\ref{fig:reel_baseline}, \ref{fig:color_issue}, and \ref{fig:corner_artifact} is produced by Netflix, with \texttt{CC BY-NC-ND 4.0} license:\\ {\small\texttt{https://media.xiph.org/video/derf/ElFuente/Netflix\_Tango\_Copyright.txt}}

\section{\texttt{ffmpeg} and \texttt{HM} experiments}
\label{sec:appendix_baseline_generation}

We generated H.264 and H.265 baselines using \texttt{ffmpeg} and \texttt{HM} software tools. Both \texttt{HM} and \texttt{ffmpeg} received the UVG videos in the native YUV-1080p-8bit format as inputs. The command that we used to run \texttt{ffmpeg} in low-latency mode is as follows:

\begin{quote}
\noindent\small \texttt{ffmpeg -y -pix\textunderscore fmt yuv420p -s [W]x[H] -r [FR] -i [IN].yuv \\
-c:v libx[ENC] -b:v [RATE]M -maxrate [RATE]M -tune zerolatency \\
-x[ENC]-params "keyint=[GOP]:min-keyint=[GOP]:verbose=1" [OUT].mkv}
\end{quote}
 
\noindent and the command that we used to run \texttt{ffmpeg} in default settings is as follows:

\begin{quote}
\noindent\small  \texttt{ffmpeg -y -pix\textunderscore fmt yuv420p -s [W]x[H] -r [FR] -i [IN].yuv \\
-c:v libx[ENC] -b:v [RATE]M -maxrate [RATE]M \\
-x[ENC]-params "verbose=1" [OUT].mkv}
\end{quote}

\noindent where the values in brackets represent the encoder parameters as follows: \texttt{H} and \texttt{W} are the frame dimensions ($1080\times1920$), \texttt{FR} is the frame rate (120), \texttt{ENC} is the encoder type (x264 or x265), \texttt{GOP} is the GoP size (12 for low-latency settings),  \texttt{INPUT} and \texttt{OUTPUT} are the input and the output filenames, respectively, \texttt{RATE} controls the intended bit rate in Mega-bits/second (We tried \texttt{RATE}$=\{10, 20, 37, 62, 87, 112\}$ Mega-bits/second that translate to $\{0.04, 0.08, 0.15, 0.25, \\0.35, 0.45\}$ bits/pixel for UVG 1080p at 120 fps). It is worth to mention that we did not set the \texttt{-preset} flag in \texttt{ffmpeg} that controls the coding performance. As a result, it used the default value which is \texttt{medium}, unlike some existing papers that employed \texttt{ffmpeg} in \texttt{fast} or \texttt{superfast} presets that led to poor \texttt{ffmpeg} performance.

The command we used to run \texttt{HM} is as follows:

\begin{quote}
\noindent \small \texttt{TAppEncoderStatic -c [CONFIG].cfg -i INPUT.yuv -wdt [W] -hgt [H] \\
-fr [FR] -f [LEN] -o OUTPUT.yuv -b -ip [GOP] -q [QP] > [LOG].log}
\end{quote}

\noindent where \texttt{CONFIG} is the \texttt{HM} configuration file (we used \texttt{LowDelayP.cfg}), \texttt{LEN} is the number of frames in the video sequence (600 or 300 for UVG videos), \texttt{LOG} is the log file name that we used to read the bit rate values, and \texttt{QP} control the quality of the encoded video that leads to different bit rates (we tried \texttt{QP}$=\{20, 22, 25, 30, 35, 40\}$ in this work).

MS-SSIM for all the experiments was calculated in RGB domain in video-level. bit rates for \texttt{ffmpeg} were calculated by reading the compressed file size and for \texttt{HM} by parsing the \texttt{HM} log file. Both values were then normalized by frame-dimensions and frame-rate. 

\subsection{YUV to RGB conversion inconsistencies}
\begin{figure}
 \centering
 \includegraphics[height=6.5cm]{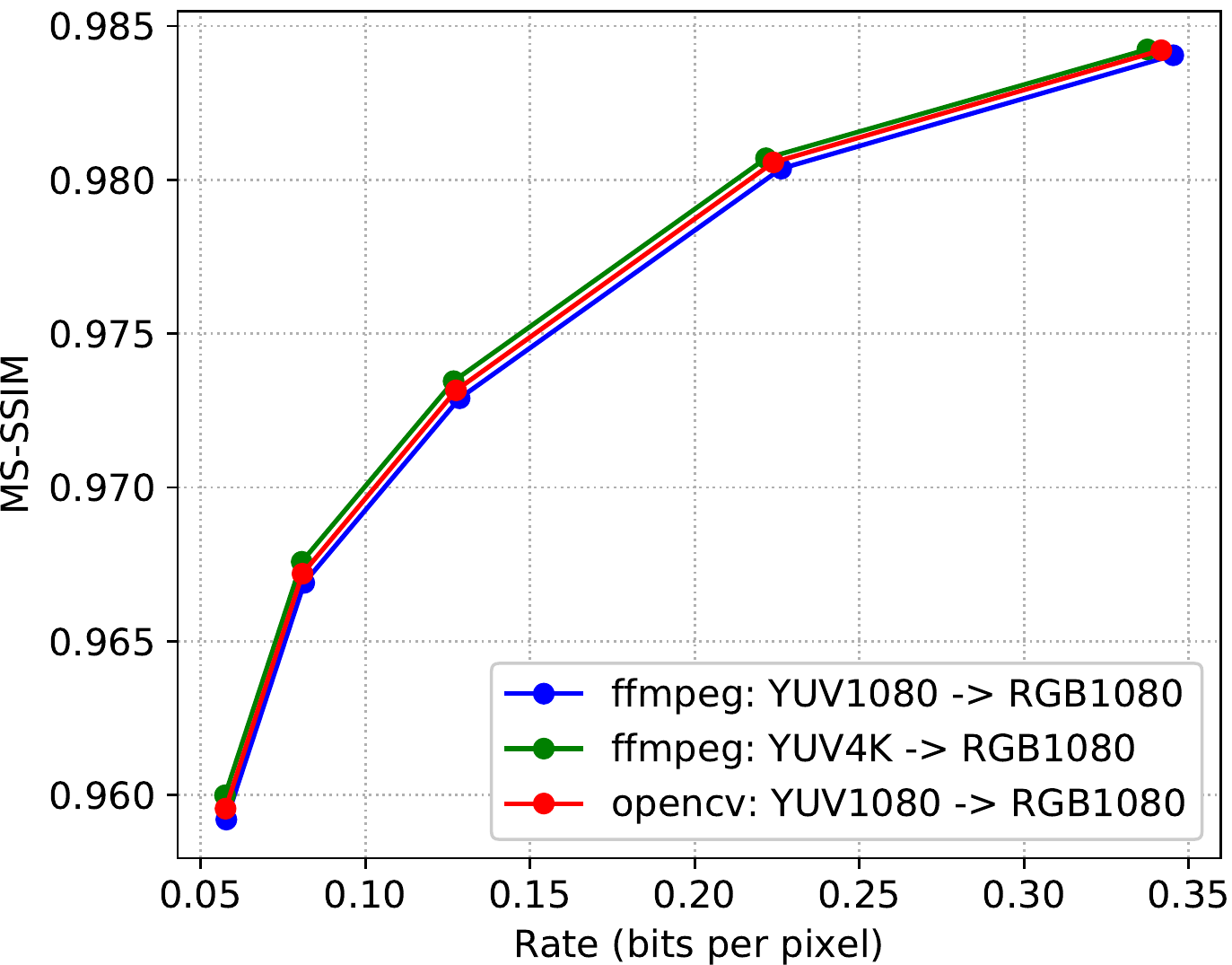}
 \caption{Effect of conversion method on model performance on UVG dataset.}
 \vspace{-0.5cm}
 \label{fig:uvg_versions}
\end{figure}

Since the UVG raw frames are available in YUV format and our model works in RGB domain only, we had to convert the UVG frames from YUV to RGB color space. We considered two methods to convert UVG-1080p-8bit videos to RGB $1080p$, \emph{i)} use \texttt{ffmpeg} to read in YUV format and directly save the frames in RGB format, \emph{ii)} read the frames in YUV and convert them to RGB via \texttt{COLOR\textunderscore YUV2BGR\textunderscore I420} functionality provided by the \texttt{OpenCV}~\cite{opencv} package. We also tried a third scenario, \emph{iii)} use \texttt{ffmpeg} to read UVG-4K-10bit in YUV format and save the RGB $1080p$ frames. Fig.~\ref{fig:uvg_versions} shows the performance of our model on the three different versions of UVG RGB. As can be seen from this figure, there is an inconsistency in our model performance across different UVG versions. A potential cause for this behavior could be the interpolation technique employed in the color conversion method, since in YUV420 format U and V channels are subsampled and need to be upsampled before YUV to RGB conversion. 
The results reported in this paper are based on the \texttt{OpenCV} color conversion.

\section{Qualitative examples}\label{appendix:qualitative_examples}

\begin{sidewaysfigure}
 \centering
 \includegraphics[width=0.90\textwidth]{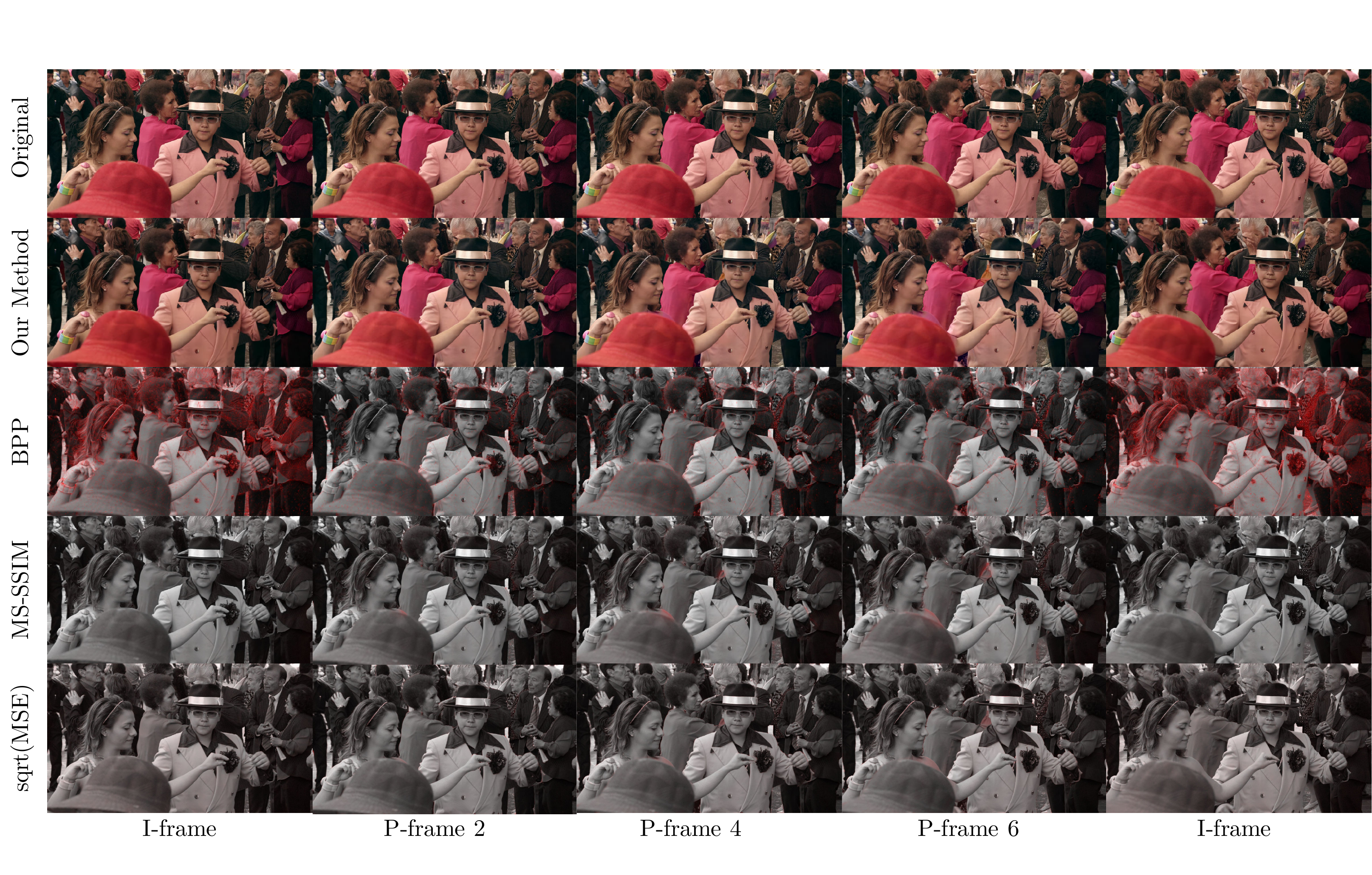}
 \caption{Our method's qualitative evaluation on frames from Netflix Tango in Netflix El Fuente. The rows show the original frames, the reconstructed frames, then the BPP, MS-SSSIM and MSE maps overlayed on top of the grayscale original frames. Each column shows a different frame from a full cycle of GoP 8 going from I-frame to I-frame, showing every other frame.}
 \vspace{-0.5cm}
 \label{fig:reel_baseline}
\end{sidewaysfigure}

\begin{sidewaysfigure}
 \centering
 \includegraphics[width=0.90\textwidth]{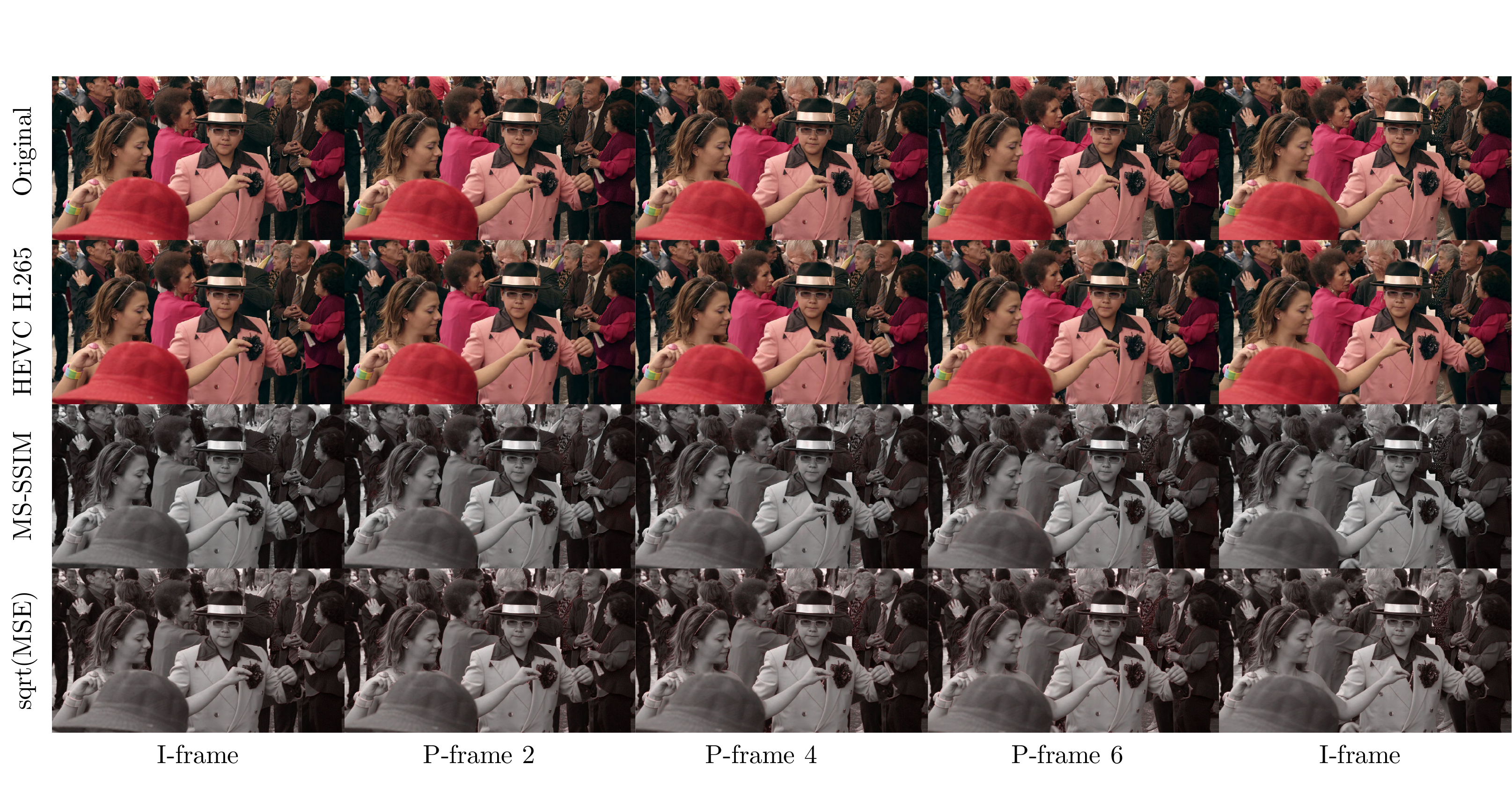}
 \caption{H.265 qualitative evaluation on frames from Netflix Tango in Netflix El Fuente. The rows show the original frames, the reconstructed frames, then the MS-SSSIM and MSE maps overlayed on top of the grayscale original frames. Each column shows a different frame from a full cycle of GoP 8 going from I-frame to I-frame, showing every other frame.}
 \vspace{-0.5cm}
 \label{fig:reel_hevc}
\end{sidewaysfigure}

We provide in this section more qualitative examples for our method in Fig.~\ref{fig:reel_baseline} and for H.265 in Fig.~\ref{fig:reel_hevc}. Notably, we overlay the MS-SSIM and MSE error maps on the original frame to provide indication of the distortion characteristics of each algorithm.
In addition in Fig.~\ref{fig:reel_baseline}, we show the BPP maps, as we have access to the entropy of the corresponding latent using the I-frame and P-frame prior models. We average the entropy across channels, then use bilinear upsampling to bring the BPP map back to the original image resolution.

By examining the BPP maps in Fig.~\ref{fig:reel_baseline} (third row), we can see that intuitively the model spends bits more uniformly across I-frames (first and last column), while during P-frames bits are mostly spent around edges of moving objects.

Note that our method optimizes for MS-SSIM, while traditional video codecs like H.265 are 
evaluated with 
PSNR. As detailed in Appendix~\ref{appendix:msssim_color_issue}, MS-SSIM is
insensitive 
to uniform color shifts and mainly focuses on spatial structure. On the contrary, PSNR does pick up color shifts, but
it tends to care less for fine details in spatial structure, \ie texture, sharp edges. Our method's MS-SSIM maps (Fig.~\ref{fig:reel_baseline} 4th row) show edges are sharply reconstructed, yet does not fully capture some of the color shifts in the reconstruction. In particular the man in the pink suit has a consistently dimmer pink suit in the reconstruction (compare first and second row), which does not appear in the MS-SSIM loss, while MSE seems to pick up the shift. In contrast in Fig.~\ref{fig:reel_hevc}, H.265 MSE maps (last row) shows that the error is concentrated around most edges and fine textures due to blurriness, yet most colors are on target.

We observe that for our method, MS-SSIM error maps degrade as we go from I-frame to P-frame (compare first and second to last column in Fig.~\ref{fig:reel_baseline}). Similarly, the BPP map of the first I-frame show a large rate expenditure, while consecutive P-frames display significantly less rate. Both of these qualitative results confirm the quantitative analysis of Fig.~\ref{fig:bpp_msssim_per_frame}.

\section{Theoretical justification of the feedback recurrent module}
\label{appendix:theoretical_justification}
 
In this subsection we provide theoretical motivations on 
having feedback recurrency module in our network design.
Specifically, we show that, for the compression of sequential data with a causal decoder and causal encoder, (i) it is imperative for 
encoder to have a memory that summarizes previously latent codes, and (ii) it is not beneficial for encoder to access history input information.

Below is
an abstract view of a generic sequential autoencoder, where $\mathbf{X}=\{\v{x}_\tau\}_{\tau\in\mathbb{N}}$ denotes a discrete-time random process that represents the empirical distribution of our training video data, with $\mathbf{Z}$ and $\mathbf{\widehat{X}}$ being the induced latent and reconstruction processes.
\begin{align}
    \mathbf{X} \xrightarrow{f_\text{enc}} \mathbf{Z} \xrightarrow{f_\text{dec}} \mathbf{\widehat{X}}. \notag
\end{align}
From an information theoretical point of view, the sequential autoencoding problem can be abstracted as the maximization of average frame-wise mutual information between $\mathbf{X}$ and $\mathbf{\widehat{X}}$, 
 detailed below, with certain rate constraint on $H(\mathbf{Z})$,
\begin{align}
    \underset{f_\text{enc}, f_\text{dec}}{\text{max.}}\text{ }& \textstyle\sum_{\tau}\nolimits I\left(\v{x}_\tau; \widehat{\v{x}}_\tau\right),&\text{frame-wise mutual information\footnotemark}\notag\\
    \text{s.t. }
    & I\left(\v{x}_\tau;\widehat{\v{x}}_\tau|\v{z}_{\leq \tau}\right)=0,\text{ }\forall \tau.&\text{decoder causality}\notag
\end{align}
The decoder causality is encoded in such a form as $I(\v{x}_\tau;\widehat{\v{x}}_\tau$ $|\v{z}_{\leq \tau})$ is zero \emph{if and only if} $\widehat{\v{x}}_\tau$ is not a function of $\v{z}_{>\tau}$. It is important to note that mutual information is invariant to any bijections and thus would not reflect perceptual quality of reconstruction. Nevertheless, we use this formulation only to figure out important data dependencies from an information theory perspective and use that to guide us in the design of network architecture. With a closer look, the $t^\text{th}$ term in the objective function can be rewritten as below. \footnotetext{Note that this is different from $I(\mathbf{X};\mathbf{\widehat{X}})$. If we instead maximize $I(\mathbf{X};\mathbf{\widehat{X}})$, then we could have a single output frame capturing the information of more than one input frames, which is not what we want.
}
\begin{align}
& I (\v{x}_t;\widehat{\v{x}}_t)\notag\\
=&I(\v{x}_t; \widehat{\v{x}}_t, \v{z}_{\leq t}) - I(\v{x}_t; \v{z}_{\leq t}|\widehat{\v{x}}_t)\notag\\
=&I(\v{x}_t;\v{z}_{\leq t}) + I(\v{x}_t;\widehat{\v{x}}_t|\v{z}_{\leq t}) - I(\v{x}_t;\v{z}_{\leq t}|\widehat{\v{x}}_t)  \notag\\
(a)=&I(\v{x}_t;\v{z}_{\leq t})  - I(\v{x}_t;\v{z}_{\leq t}|\widehat{\v{x}}_t) \notag\\
=&I(\v{x}_t;\v{z}_{<t})+I(\v{x}_t; \v{z}_t|\v{z}_{<t}) - \left(
I(\v{x}_t; \v{z}_{<t}|\widehat{\v{x}}_t)+
I(\v{x}_t; \v{z}_t|\widehat{\v{x}}_t, \v{z}_{<t})
\right)\notag\\
(b)=&\underbrace{I(\v{x}_t;\v{z}_{<t};\widehat{\v{x}}_t)}_{\text{prediction}} + \underbrace{I(\v{x}_t;\v{z}_t;\widehat{\v{x}}_t|\v{z}_{<t})}_{\text{innovation}}.\label{eq:mutual_information_break_down}
\end{align}
Step (a) incorporates the decoder causality constraint, step (b) comes from the definition of multi-variate mutual information\footnote{$I(a;b;c)=I(a;b)-I(a;b|c)$. For a Markov process $a\to b\to c$, this indicates the amount of information flown from $a$ to $c$ through $b$.}, and the rest uses the identity of conditional mutual information\footnote{$I(a;b)=I(a;b,c)-I(a;c|b)$ for any $c$.}. 

Equation~\eqref{eq:mutual_information_break_down} says that $I(\v{x}_t;\widehat{\v{x}}_t)$ can be broken down into two terms, the first represents the prediction of $\v{x}_t$ from all previous latents $\v{z}_{<t}$, and the second represents new information in $\v{z}_t$ about $\v{x}_t$ that is not be explained by $\v{z}_{<t}$.
While this formulation does not indicate how the optimization can be done, it tells us what variables should be incorporated in the design of $f_\text{enc}$. Let us focus on $\v{z}_t$: in our objective function $\textstyle\sum_{\tau}\nolimits I(\v{x}_\tau;\widehat{\v{x}}_{\tau})$, $\v{z}_t$ shows up in the following terms
\begin{align}
I(\v{x}_\tau; \v{z}_\tau; \widehat{\v{x}}_\tau|\v{z}_{<\tau})\text{ for }\tau\geq t \text{, and } I(\v{x}_\tau;\v{z}_{<\tau};\widehat{\v{x}}_\tau) \text{ for }\tau>t.\notag
\end{align}
It is clear, then, that the optimal $\v{z}_t$ should only be a function of $\v{x}_{\geq t}$, $\widehat{\v{x}}_{\geq t}$, $\v{z}_{<t}$ and $\v{z}_{>t}$. Combining it with the constraint that the encoder is causal, we can further limit the dependency to $\v{x}_t$ and $\v{z}_{<t}$. In other words, it suffices to parameterize the encoder function as $\v{z}_t=f_\text{enc}(\v{x}_t, \v{z}_{<t})$, which attests to the two claims at the beginning of this subsection: (i) $\v{z}_t$ should be a function of $\v{z}_{<t}$ and (ii) $\v{z}_t$ does not need to depend on $\v{x}_{<t}$.

It is with these two claims that we designed the network where the decoder recurrent state, which provides a summary of $\v{z}_{<t}$, is fed back\footnote{Since $\widehat{x}_{<t}$ is a deterministic function of $\v{z}_{<t}$, any additional input of $\widehat{x}_{<t}$ to the encoder is also justified.} to the encoder as input at time step $t$ and there is no additional input related to $\v{x}_{<t}$.

It is worth noting that in \cite{Gregor2015,Gregor2016}, the authors introduce a neural network architecture for progressive coding of images, and in it an \emph{encoder recurrent connection} is added on top of the feedback recurrency connection from the decoder. Based on the analysis in this section, 
since the optimization of $\v{z}_t$ does not depend on $\v{x}_{<t}$, we do not include encoder recurrency in our network design.

\section{Graphical modeling considerations}
\label{appendix:graphical_modelling_considerations}

In this section we give more details on the reasoning behind how we formulated the graphical models presented in Fig.~\ref{fig:graphic_model_related_work}.

\vspace{5pt}
\noindent\textbf{Wu \etal\cite{wuVideoCompressionImage2018}} \\
\noindent\emph{Generative model.}\text{     }
Temporally independent prior is used hence no edges between the latent variables $\v{z}_t$. 
Consecutive reconstructions $\widehat{\v{x}}_t$ depend on the previous reconstructions $\widehat{\v{x}}_{t-1}$ and hence the edge between them $\widehat{\v{x}}_{t-1} \rightarrow \widehat{\v{x}}_t$.
\\
\noindent\emph{Inference model.}\text{     }
During inference, 
at timestep $t$, 
the current timestep's latent $\v{z}_{t}$ is inferred based on the previous reconstruction $\widehat{\v{x}}_{t-1}$ and current original frame $\v{x}_t$.
In turn, the previous reconstruction $\widehat{\v{x}}_{t-1}$ is determined based on the information from the previous timestep's latent $\v{z}_{t-1}$ and the earlier reconstruction $\widehat{\v{x}}_{t-2}$.
Since the procedure of determining $\widehat{\v{x}}_{t-1}$ from $\v{z}_{t-1}$ and $\widehat{\v{x}}_{t-2}$ 
is deterministic, the inference model has an edge $\v{z}_{t-1} \rightarrow \v{z}_{t}$.
Extending this point, since $\widehat{\v{x}}_{t}$ is updated at each timestep using a deterministic procedure using its previous value $\widehat{\v{x}}_{t-1}$ and $\v{z}_{t}$, it makes $\v{z}_{t}$ dependent on all the past latent codes $\v{z}_{<t}$.
Hence there are edges from all the past latents $\v{z}_{<t}$ to the present latent $\v{z}_{t}$. \\
The edge $\v{x}_{t} \rightarrow \v{z}_{t}$ arises from the direct dependence of $\v{z}_{t}$ on $\v{x}_{t}$ through the encoder.

\vspace{5pt}
\noindent\textbf{Liu \etal \cite{liu2020learned} } \\
\noindent\emph{Generative model.}\text{     }
Due to the introduction of the time-autoregressive code prior there are additional edges $\v{z}_{<t} \rightarrow \v{z}_{t}$ as compared to model (a) with dashed edges.
\\
\noindent\emph{Inference model.}\text{     }
The inference model remains unchanged.

\begin{figure}[hbt!]
 \centering
 \includegraphics[width=0.7\linewidth]{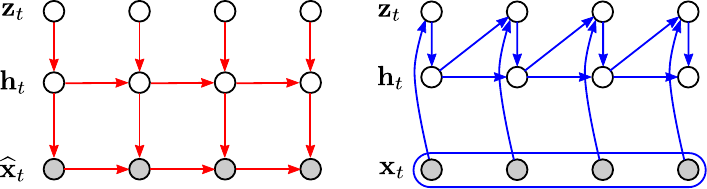}
 \caption{Left: Generative model of our method with intermediate variable $\v{h}_t$. 
 Right: Inference model of our method with intermediate variable $\v{h}_t$.}
 \label{fig:model_b_with_hiddens}
 \end{figure}

\vspace{5pt}
\noindent\textbf{Rippel \etal \cite{rippelLearnedVideoCompression2018} and ours} \\
\noindent\emph{Generative model.}\text{     }
Model (b) differs from model (a) in the introduction of the recurrent connection and hence hidden state $\v{h}_t$, hence the full graphical model could be drawn as per Fig.~\ref{fig:model_b_with_hiddens}.
However, the $\v{h}_t$ is not a stochastic random variable, but a result of a deterministic function of $\v{h}_{t-1}$ and $\v{z}_t$.
For that reason we can drop explicitly drawing nodes $\v{h}_t$ and move the deterministic relationship they are implying into the graphical model edges between the latent variables $\v{z}_t$.
This way we arrive from the models in Fig.~\ref{fig:model_b_with_hiddens} to models in Fig.~\ref{fig:graphic_model_related_work}(b).
\\
\noindent\emph{Inference model.}\text{     }
Analogous reasoning applies in the case of the inference model.

\noindent\textbf{Han \etal \cite{stephan_2019_neurips}} \\
Graphical models are derived based on equations by Han \etal \cite{stephan_2019_neurips}: Eq.~(4) and Eq.~(6) for the generative model, and Eq.~(5) for the inference model.

\vspace{5pt}
\noindent\textbf{Habiban \etal \cite{Habibian_2019_ICCV}} \\
\noindent\emph{Generative model.}\text{     }
Due to the use of a prior which is autoregressive in the temporal dimension (across the blocks of 8 frames that Habiban \etal are using) we draw edges $\v{z}_{<t} \rightarrow \v{z}_{t}$.
The latent codes are generated by a deterministic composition of 3D convolutional layers with the input of the original frames. The temporal receptive field is larger than original input sequence which means that every latent variable in the block can be influenced by each of the original input frames and hence the fully connected clique between $\v{z}_{1:T}$ and $\v{x}_{1:T}$.
\\
\noindent\emph{Inference model.}\text{     }
The same reasoning applies for the inference model.

\subsection{Full flexibility of the marginal $\mathbb{P}_\v{X}(\v{x}_{1:T})$}
\label{appendix:full_flexibility_marginal}
The graphical model corresponding to the approach of Liu \etal \cite{liu2020learned} is presented in Fig.~\ref{fig:graphic_model_related_work}(a) including dashed lines.  
In this case, the marginal $\mathbb{P}_\v{X}(\v{x}_{1:T})$ is fully flexible in the sense that it does not make any conditional independence assumptions between $\v{x}_{1:T}$.

To see that consider considering what happens when we marginalize out $\v{z}_{1:T}$ from 
$\mathbb{P}_{\v{Z}}(\v{z}_{1:T})\mathbb{P}_{\v{X}|\v{Z}}(\v{x}_{1:T}|\v{z}_{1:T})$ 
using the variable elimination algorithm \cite{koller2009probabilistic}.
Assume we use an elimination order $\tau$ which is permutation of a set of numbers $\{1, \dots, T\}$, e.g. for $T=4$ a viable permutation is $\tau=\{2,1,4,3\}$.
Let's think about the process of variable elimination in terms of the induced graph.
Eliminating the first variable $\v{z}_{\tau_1}$ induces a connection between $\v{x}_{\tau_1}$ and each of the other latent variables $\v{z}_{\neq \tau_1}$.
Eliminating consecutive latent variables $\v{z}_{t}$ will induce connections between corresponding observed variables $\v{x}_{t}$ and every other $\v{z}_{\neq t}$.
This way the final latent variable to be eliminated $\v{z}_{\tau_T}$ will be connected in the induced graph with all the observed variables $\v{x}_{1:T}$.
Hence when we eliminate $\v{z}_{\tau_T}$ that will result in a factor $\phi(\v{x}_{1:T})$ (a clique between all nodes $\v{x}_{1:T}$ thinking in terms 2of the induced graph).
The marginal distribution we are looking for is $\mathbb{P}_\v{X}(\v{x}_{1:T}) \propto \phi(\v{x}_{1:T})$,
and it makes no conditional independence assumptions.

The graphical model corresponding to our and Rippel \etal approach is presented in Fig.~\ref{fig:graphic_model_related_work}(b).
In this case, showing that the marginal $\mathbb{P}_\v{X}(\v{x}_{1:T})$ makes no conditional independence assumptions between $\v{x}_{1:T}$ is simpler -- 
no matter what variable elimination order we choose, since $\v{z}_1$ is connected to all the nodes $\v{x}_{1:T}$ eliminating $\v{z}_1$ will always result in a factor $\phi(\v{x}_{1:T})$.

\section{MS-SSIM color issue and corner artifact}
\label{appendix:msssim_color_issue}

\subsection{MS-SSIM color shift artifact}

MS-SSIM, introduced in \cite{wangMultiScaleStructuralSimilarity2003}, is known to be to some extent invariant to color shift \cite{l1_plus_msssim}. Inspired by the study in Fig.~4 of \cite{Wang2009MSELoveItLeaveIt}, we designed a small experiment to study the extent of this issue. We disturbed an image $\v{x}$ slightly with additive Gaussian noise to obtain $\widehat{\v{x}}_0$, such that $\widehat{\v{x}}_0$ lies on a certain equal-MS-SSIM hypersphere (we choose 0.966 which corresponds to the average quality of our second lowest rate model). Then we optimize $\widehat{\v{x}}$ for lower PSNR, under the constraint of equal MS-SSIM 
using the following objective:

\begin{align}
    \mathcal{L} = \text{PSNR}(\v{x}, \hat{\v{x}}) + \alpha_t \cdot \left|\text{MS-SSIM}(\v{x}, \hat{\v{x}}) - \text{MS-SSIM}(\v{x}, \hat{\v{x}}_0)\right|
\end{align}

We use Adam optimizer with a learning rate of $10^{-4}$ (otherwise default parameters) for $T=20,000$ iterations. We use $\alpha_t = \min(1, \frac{2t}{T} )\times 10^3$, with $t$ representing the iteration index.

\begin{figure}
\centering
\includegraphics[width=.99\linewidth]{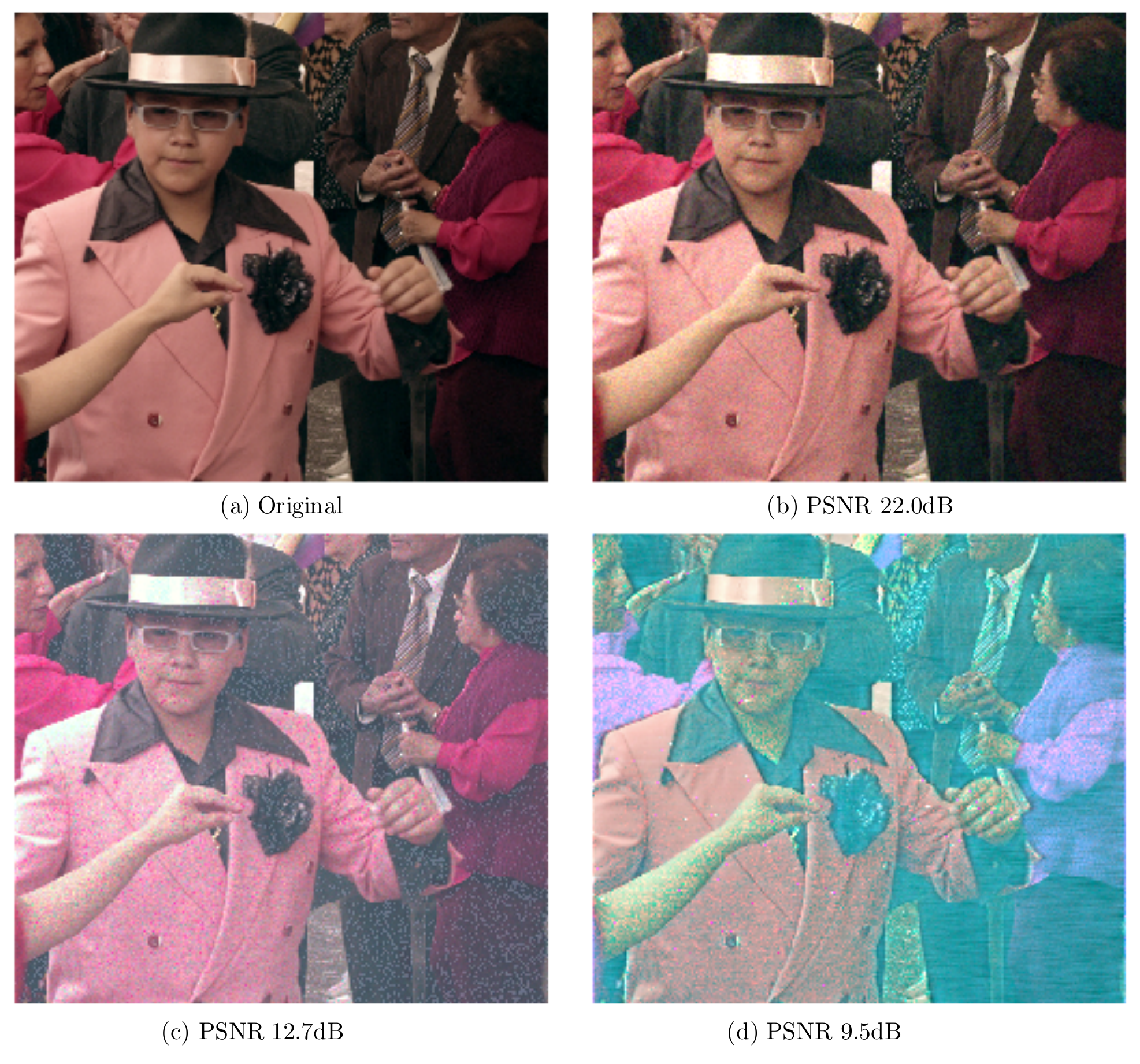}
\caption{An illustration of MS-SSIM color issue. All frames (b), (c) and (d) have an MS-SSIM of 0.966. (a): original frame, (b): perturbed image with slight additive gaussian noise, (c) and (d): images obtained by optimizing (b) for $\mathcal{L}=\text{PSNR}$ under equal MS-SSIM contraint.\\ Crop of frame 229 of Tango video from Netflix Tango in Netflix El Fuente; see Appendix~\ref{appendix:image_attribution} for license information.
}
\label{fig:color_issue}
\end{figure}

In Fig.~\ref{fig:color_issue}, (a) corresponds to $\v{x}$, (b) corresponds to $\hat{\v{x}}_0$, (c) to $\hat{\v{x}}$ at iteration $2,000$ and (d) to $\hat{\v{x}}$ at iteration $20,000$. All frame (b), (c) and (d) are on the equal-MS-SSSIM hypersphere of 0.966 with respect to the original frame $\v{x}$. We can observe that it is possible to obtain an image with very low PSNR, mainly due to drastic uniform color shifts, while remaining at equal distance in terms of MS-SSIM from the original image. This helps explain some of the color artifacts we have seen in the output of our models.

\subsection{MS-SSIM corner artifact}

\begin{figure}
\centering
\includegraphics[width=.99\linewidth]{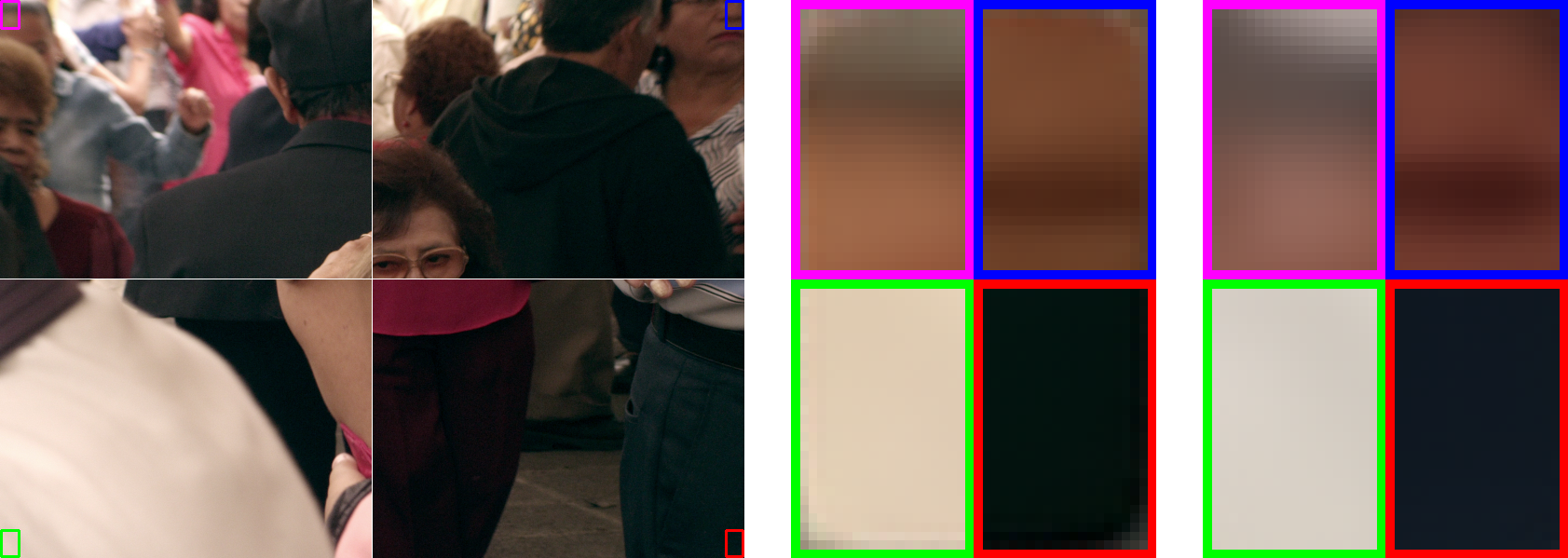}
\caption{An illustration of the MS-SSIM corner artifact, best viewed on screen. Left: uncompressed frame, middle: corners when trained with default MS-SSIM implementation, right. corners when trained with replicate-padded MS-SSIM implementation.\\ Frame 10 of Tango video from Netflix Tango in Netflix El Fuente; see Appendix~\ref{appendix:image_attribution} for license information.
}
\label{fig:corner_artifact}
\end{figure}

The standard implementation of MS-SSIM, when used in $\mathcal{L}_\text{RD}$, causes corner artifacts as shown in Figure~\ref{fig:corner_artifact}. Although the artifact is subtle and barely noticeable when looking at full-size video frames, it is worth investigation. We found the root cause in the convolution-based implementation of local mean calculations in SSIM function where a Gaussian kernel (normally of size 11) is convolved with the input image. The convolution is done without padding, as a result the contribution of the corner pixels to MS-SSIM is minimal as they fall at the tails of the Gaussian kernel. This fact is not problematic when MS-SSIM is used for evaluation, but can negatively affect training. Specifically in our case, where MS-SSIM was utilized in conjunction with rate to train the network, the network took advantage of this deficiency and assigned less bits to the corner pixels as they are less important to the distortion term. As a result, the corner pixels appeared blank in the decoded frames. We fixed this issue by padding, \eg replicate, the image before convolving it with the Gaussian kernel. The artifact was eliminated as can be seen from Fig.~\ref{fig:corner_artifact}.

It is worth to mention that we used the standard MS-SSIM implementation in all the trainings and evaluations in this paper for consistency with other works.

\end{document}